\documentclass{article} 
\usepackage{iclr2027_conference,times}

\usepackage{etoolbox}
\makeatletter
\patchcmd{\@maketitle}
  {\lhead{Published as a conference paper at ICLR 2027}}
  {\lhead{}}
  {}
  {\PackageError{arxiv-preprint}{Could not remove conference publication header}
    {Check the supplied ICLR style.}}
\makeatother
\iclrfinaltrue
\renewcommand{\headrulewidth}{0pt}

\usepackage{url}

\usepackage[utf8]{inputenc}
\usepackage[T1]{fontenc}
\usepackage{booktabs}
\usepackage{amsfonts}
\usepackage{amsmath}
\usepackage{nicefrac}
\usepackage{microtype}
\usepackage{xcolor}
\usepackage{siunitx}
\usepackage{graphicx}
\usepackage{float}
\usepackage{hyperref}
\usepackage[capitalize,noabbrev]{cleveref} 
\hypersetup{colorlinks=true, linkcolor=blue!55!black, citecolor=teal!50!black, urlcolor=blue!55!black} 
\hypersetup{pdftitle={When Instructions Retrieve Trajectories: Diagnosing and Mitigating Generalization Failures in VLA Models}}
\hypersetup{pdfauthor={Hung-Jen Chen, Yu-Hsun Hou, Yan-Hong Chen, Yan-Fu Chen, Binghua Cai, Min Sun, Chun-Yi Lee}}
\usepackage{multirow}
\usepackage{enumitem}
\usepackage[normalem]{ulem}
\usepackage{pifont}

\setlist{nosep,leftmargin=*}
\usepackage{xspace}

\DeclareMathOperator*{\argmin}{arg\,min}

\AtBeginDocument{%
  \crefname{figure}{Fig.}{Figs.}%
  \Crefname{figure}{Fig.}{Figs.}%
  \crefname{table}{Table}{Tables}%
  \Crefname{table}{Table}{Tables}%
  \crefname{section}{Section}{Sections}%
  \Crefname{section}{Section}{Sections}%
  \crefformat{equation}{Eq.~(#2#1#3)}%
  \Crefformat{equation}{Eq.~(#2#1#3)}%
  \crefrangeformat{equation}{Eqs.~(#3#1#4) to~(#5#2#6)}%
  \crefmultiformat{equation}{Eqs.~(#2#1#3)}{ and~(#2#1#3)}{, (#2#1#3)}{ and~(#2#1#3)}%
  \Crefmultiformat{equation}{Eqs.~(#2#1#3)}{ and~(#2#1#3)}{, (#2#1#3)}{ and~(#2#1#3)}%
}

\usepackage{colortbl}

\newcommand{\bp}[1]{{\scriptsize\textcolor{teal}{(+#1)}}}
\newcommand{\bn}[1]{{\scriptsize\textcolor{gray!80!black}{($-$#1)}}}  
\newcommand{\bz}[1]{{\scriptsize\textcolor{gray!80!black}{(#1)}}}  

\title{When Instructions Retrieve Trajectories: \\Diagnosing and Mitigating \\Generalization Failures in VLA Models}

\author{%
Hung-Jen Chen\textsuperscript{1} \quad
Yu-Hsun Hou\textsuperscript{1,}\thanks{Equal contribution.} \quad
Yan-Hong Chen\textsuperscript{1,}\footnotemark[1] \quad
Yan-Fu Chen\textsuperscript{1,}\footnotemark[1] \\
\textbf{Binghua Cai\textsuperscript{1} \quad Min Sun\textsuperscript{1} \quad Chun-Yi Lee\textsuperscript{2}} \\
\normalfont\textsuperscript{1}National Tsing Hua University \quad
\textsuperscript{2}National Taiwan University \\
\normalfont\texttt{andyqmongo@gapp.nthu.edu.tw}}

\begin{document}

\maketitle

\begin{abstract}
Vision-language-action (VLA) models can exceed 90\% success on in-distribution tasks and withstand \emph{nuisance} changes that preserve the required action, yet fail under \emph{counterfactual} changes that demand a different action. Aggregate robustness scores can therefore conceal a more specific failure, in which a policy responds to both language and vision yet does not combine them to select the action the task requires. We call this failure \emph{instruction-action binding}. Instructions cue familiar trajectory families, and visual feedback adjusts their execution. Behavioral analyses of fine-tuned $\pi_{0.5}$ and GR00T-N1.7 policies reveal that failed rollouts often retain the source behavior or switch to another demonstrated task. These switches show that language is not simply ignored. Readouts and interventions connect these choices to task-conditioned internal states. Our analysis of the imitation objective shows how narrow conditional action support can leave grounded and instruction-keyed solutions indistinguishable on the demonstrations. This motivates \emph{Equivariant Counterfactual Training} (ECT), which acts at two levels. \emph{ECT data} supply valid demonstrations in which the same instruction requires different actions in distinguishable scenes, while the \emph{ECT loss} trains each demonstration with its counterpart in the same update. In a controlled LIBERO-PRO comparison, full ECT raises $\pi_{0.5}$'s mean position-swap success from 36\% to 59\%. On CALVIN, where counterparts already occur in the original data, the ECT loss improves five-task completion without new demonstrations. On a real UR5e under a fixed demonstration budget, full ECT raises unseen-position success from 8\% to 88\%.

\end{abstract}

\section{Introduction}
\label{sec:intro}
\begin{figure}[t]
    \centering
    \includegraphics[width=\linewidth]{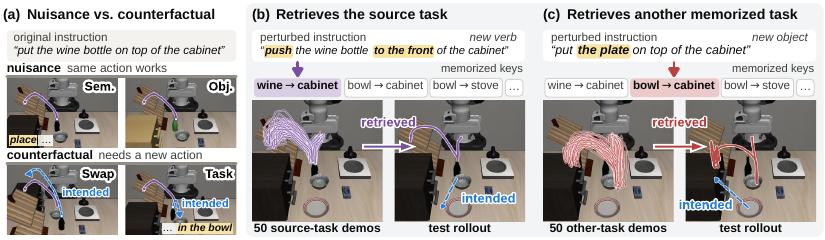}
    \caption{\textbf{Perturbed instructions retrieve memorized trajectories.} One LIBERO-Goal task throughout; dashed blue: intended trajectory, \textcolor{violet}{purple}: the source task's memorized trajectory, \textcolor{red}{red}: another memorized task's.
    \textbf{(a)} Nuisance perturbations keep the memorized trajectory valid; counterfactual ones need a new one (Sem.: rephrasing, Obj.: new appearance, Swap: swapped positions, Task: new target).
    \textbf{(b)} Under a new verb, $\pi_{0.5}$ retrieves the source task.
    \textbf{(c)} Under a new object, it retrieves another memorized task: the instruction selects which trajectory to replay.
    Of 381 labeled rollouts, 52\% replay the source task and 15\% another task (\Cref{tab:app_human_rule_counts}).}
    
    \label{fig:teaser}
\end{figure}

A robot can execute a fluent manipulation and still follow the wrong instruction. Asked to put a plate on the cabinet rather than a bowl, a fine-tuned $\pi_{0.5}$ policy can complete the familiar bowl placement instead, with a clean grasp and a trajectory resembling its demonstrations (\Cref{fig:teaser}). Yet the same policy remains successful under many changes in wording and appearance. For vision-language-action (VLA) models \citep{zitkovich2023rt,intelligence2025pi_,kim2024openvla,o2024open}, this raises a basic question. \emph{Does success under a changed input demonstrate that the policy can select a different action?}

The distinction is whether the perturbation changes the action required for success. A paraphrase or an appearance change can leave a familiar behavior valid, whereas exchanging object positions or changing the requested goal may not. We call these \emph{nuisance} and \emph{counterfactual} perturbations, respectively. On the four LIBERO-PRO suites \citep{zhou2025libero}, both $\pi_{0.5}$ and GR00T-N1.7 perform markedly worse on counterfactual cells. Reporting the two separately reveals whether the policy adapts its action selection to the current instruction--scene combination.

The failures reveal a more specific problem than ignoring language or vision. Across 381 targeted $\pi_{0.5}$ probes, 69\% of rollouts follow the source task's behavior, another demonstrated task, or a hybrid, while only 4\% collapse without a coherent trajectory. We call this \emph{instruction-action binding}. The instruction selects a familiar trajectory family, and visual feedback still adapts that family's execution. A policy can therefore remain responsive to both modalities yet fail to use the current instruction and scene together to choose the required action. The pattern also appears in GR00T-N1.7. Scene-matched readouts recover the executed task from $\pi_{0.5}$'s internal state, and interventions redirect action selection in both architectures (\Cref{sec:mechanistic_evidence}).

Why can this behavior fit the training data? In the task-specific demonstrations we study, an instruction is associated with a narrow trajectory family. An instruction-keyed solution can therefore fit the supervision without learning how the required action changes across scenes, an instance of underspecification \citep{geirhos2020shortcut,d2022underspecification}. In our fixed-step sweep from 25\% to 100\% of the original data, every Swap and Task cell remains at least 34 points below in-distribution success (\Cref{tab:sweep}). The missing distinction is in \emph{conditional action support}, where different scenes require different actions under the same instruction.

These observations and our analysis of the imitation objective suggest a training principle. Under a fixed instruction, supervision should include distinguishable scenes that require different actions, because otherwise an instruction-keyed solution remains sufficient. \emph{Equivariant Counterfactual Training} (ECT) applies this principle at two levels. ECT data add the missing scene-dependent action alternatives to the supervision, and the ECT loss uses them in optimization by training each demonstration with its counterpart in the same update. The LIBERO construction builds on action-valid geometric augmentation \citep{ameperosa2025rocoda}, guided by the scene-dependent choices missing from the original supervision. We evaluate the two components separately. At matched batch size, the ECT data account for most of the Swap gain on every LIBERO-PRO suite, and the ECT loss adds a further 2 to 5 points on Spatial, Goal, and Long over independent sampling of the same pool (\Cref{tab:ect_decomposition}), while retaining the standard per-example imitation loss.

ECT applies to constructed LIBERO counterfactuals, naturally occurring CALVIN alternatives \citep{mees2022calvin}, and physical UR5e demonstrations. On CALVIN, the ECT loss exploits counterparts within the original demonstrations, raising five-task completion from 58\% to 76\%. On the robot, unseen-position success rises from $3/40$ to $35/40$ with 400 demonstrations per main model. These settings connect ECT to constructing missing action alternatives, exploiting existing ones, and collecting them under a fixed budget.

Our central scientific contribution is the diagnosis of instruction-action binding and the supervision principle it reveals. ECT operationalizes this principle through ECT data and the ECT loss. Our contributions are threefold. \textbf{(i)}~We identify instruction-action binding in two VLA architectures, showing with behavioral probes and internal-state interventions that language sensitivity and visual tracking can coexist with incorrect task-level action selection. \textbf{(ii)}~We explain how narrow conditional action support permits instruction-keyed solutions under flow matching, which motivates same-instruction alternatives that require different actions and are distinguishable from the scene. \textbf{(iii)}~We introduce ECT data and the ECT loss, evaluate them separately in controlled comparisons, and validate ECT with constructed LIBERO counterparts, naturally occurring CALVIN counterparts, and physical UR5e demonstrations.
\section{A Framework for Instruction-Action Binding}
\label{sec:framework}
Let $\mathcal{D}$ contain demonstrations $(s_i,\ell_i,a_i)$, where $s_i$ is the observation, $\ell_i$ the instruction, and $a_i$ the expert action trajectory. In the LIBERO fine-tuning setting \citep{liu2023libero}, demonstrations under the same instruction usually follow a narrow trajectory family.

Conditional entropy $H(T\mid x)$ measures how much uncertainty about task $T$ remains after observing cue $x$. In the training data, the instruction usually identifies the task; outside Goal, the scene layout does too:
\begin{equation}
H(T\mid\ell)\approx0,\qquad H(T\mid s)\approx0.
\label{eq:task_entropy}
\end{equation}
Near-zero entropy means the cue usually identifies the familiar task. With narrow within-task trajectory families, this permits selecting a familiar behavior without resolving scene-dependent action changes. Goal's ten tasks share a scene, yet changed instructions can elicit other familiar behaviors that do not satisfy the request (\Cref{sec:human_rollout}), motivating the instruction-keyed model below.

A grounded policy resolves the instruction's referent $r(\ell)$ in the scene through $g(s,r(\ell))$,
\begin{equation}
\pi_{\mathrm{ground}}(a\mid s,\ell)
=\pi\left(a\mid g(s,r(\ell))\right).
\label{eq:grounded_policy}
\end{equation}
It changes its task-level action when the grounded requirement changes \citep{shao2021concept2robot,jang2022bc,zitkovich2023rt,kim2024openvla}. In contrast, an idealized instruction-keyed lookup selects among demonstrated trajectory families according to
\begin{equation}
\pi_{\mathrm{lookup}}(a\mid s,\ell)
=\pi\left(a\mid k(\ell)\right),
\qquad
k(\ell)=\argmin_j d_\ell\left(f_\ell(\ell),f_\ell(\ell_j)\right),
\label{eq:lookup_policy}
\end{equation}
where $f_\ell$ maps instructions to representations, $d_\ell$ measures their distance, and $k(\ell)$ identifies the nearest demonstrated instruction. Actual task-conditioned states depend on vision and language \citep{huang2025otter,haon2025mechanistic}. Under binding, local visual feedback can remain active even when the selected trajectory family does not satisfy the current instruction and scene.

\subsection{Conditional action support}
\label{sec:conditional_support}
Write $\operatorname{supp}_{\mathcal{D}}(A\mid\ell)$ for the expert trajectories represented under instruction $\ell$. More demonstrations can increase sample count without adding a different action family. Wherever grounded and instruction-keyed solutions both fit the demonstrations, such supervision leaves them indistinguishable \citep{d2022underspecification}. For flow matching, let $a_0$ denote noise and $a_t=(1-t)a_0+ta$ the interpolated action at time $t$. The squared imitation loss has the population-optimal velocity
\[
v^*(a_t,t\mid s,\ell)=\mathbb{E}[a-a_0\mid a_t,t,s,\ell].
\]
If the instruction key nearly determines the demonstrated trajectory, the conditional target changes little when $(s,\ell)$ is replaced by $k(\ell)$. An instruction-keyed field can therefore fit the supervision without resolving the current scene. Both $\pi_{0.5}$ and GR00T-N1.7 use flow matching. \Cref{app:flow_matching_lookup} gives the derivation under the stated data condition.

ECT supplies valid alternatives $(s,\ell,a)$ and $(s',\ell,a')$ under a fixed instruction, with observations that distinguish which action is required. The purpose is scene-dependent action choice rather than arbitrary trajectory variation. An instruction-only model may represent multiple action modes, but cannot select between different scene-conditioned expert distributions using the instruction alone. The alternatives provide supervision for this selection.

\paragraph{Perturbation cells.}
A perturbation from $(s,\ell)$ to $(s',\ell')$ can change the selected trajectory family, the action required for success, or both. Let $a^*(s,\ell)$ denote the task-relevant action requirement, including the target, goal, and their spatial realization, while abstracting from incidental execution variation. Using the instruction-keyed model in \Cref{eq:lookup_policy}, we distinguish
\begin{equation}
\Delta_k=\mathbf{1}\!\left[k(\ell')\neq k(\ell)\right],
\qquad
\Delta_a=\mathbf{1}\!\left[a^*(s',\ell')\neq a^*(s,\ell)\right].
\label{eq:key_trajectory_delta}
\end{equation}
The two indicators separate what the policy selects from what the task requires. Nuisance changes preserve the action requirement, giving $\Delta_a=0$, whereas counterfactual changes give $\Delta_a=1$. In this model, a Swap preserves the instruction key but changes the required action. A changed instruction can either retain the source key ($\Delta_k=0$) or select another familiar task ($\Delta_k=1$). This distinction connects the source-task and other-task retrievals in \Cref{fig:app_all_rollouts} without treating a key change as correct task following. \Cref{tab:key_trajectory_decomposition} summarizes these cases. We also report in-distribution success (ID). Environment is omitted because its assets are unavailable (\Cref{app:libero_pro_patch}).

\begin{table}[t]
\centering
\scriptsize
\setlength{\tabcolsep}{3pt}
\vspace{-1.0em}
\caption{\textbf{Nuisance and counterfactual cells under the binding model.} $\Delta_k$ describes a change in the selected instruction key and $\Delta_a$ a change in the required action (\Cref{eq:key_trajectory_delta}). The Semantic row represents key-preserving paraphrases. A Task change can retain or switch the key.}
\label{tab:key_trajectory_decomposition}
\begin{tabular*}{\linewidth}{@{\extracolsep{\fill}}lcccl@{}}
\toprule
Cell & $\Delta_k$ & $\Delta_a$ & Type & Binding pattern \\
\midrule
Semantic (Sem.) & 0 & 0 & Nuisance & familiar action remains valid \\
Object (Obj.) & 0 & 0 & Nuisance & familiar action can remain valid \\
Swap & 0 & 1 & Counterfactual & same key, different required action \\
Task & 0 or 1 & 1 & Counterfactual & source retention or other-task retrieval \\
\bottomrule
\end{tabular*}
\end{table}

\subsection{Testable implications}
\label{sec:predictions}
Binding suggests that (i) nuisance performance can remain high while (ii) counterfactual changes expose a larger gap. It also suggests (iii) failures resembling familiar demonstrated behaviors and (iv) retained local tracking after incorrect target selection (\Cref{sec:failure_analysis}). We further test (v) whether task-conditioned states influence action selection (\Cref{sec:kv_retrieval}) and (vi) whether scene-dependent action alternatives help where additional repetitions do not (\Cref{sec:ect}). \Cref{sec:ect} separately evaluates the contributions of ECT data and ECT loss.

\section{Counterfactual Failures Follow Familiar Behaviors}
\label{sec:failure_analysis}

\begin{figure}[t]
\centering
\begin{minipage}[c]{0.31\linewidth}\centering
\includegraphics[width=\linewidth]{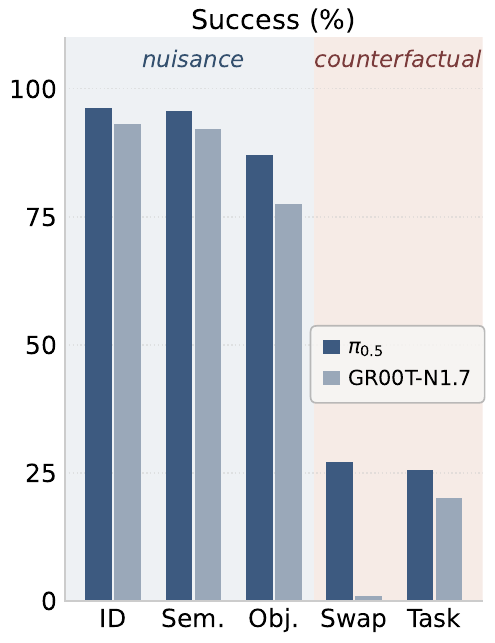}
\end{minipage}\hfill
\begin{minipage}[c]{0.67\linewidth}\centering
\includegraphics[width=\linewidth]{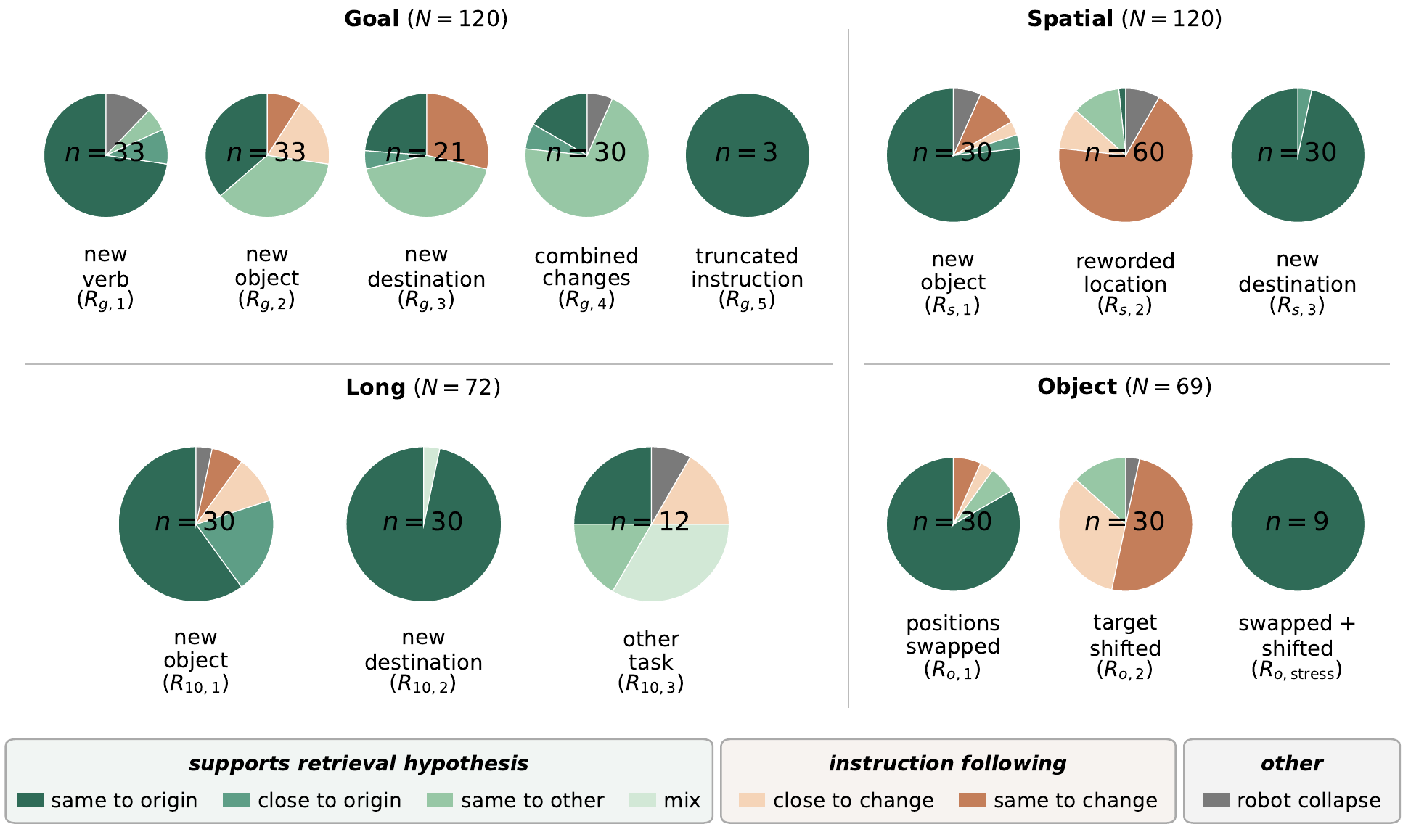}
\end{minipage}
\caption{\textbf{Counterfactual failure is often coherent rather than unstructured.}
\textbf{Left.} Official-checkpoint success on LIBERO-PRO separates nuisance and counterfactual cells.
\textbf{Right.} Each pie shows the behavioral labels for one perturbation type (see \Cref{tab:human_label_taxonomy} for label definitions). Across all 381 targeted $\pi_{0.5}$ probes, 69\% are retrieval-like and 4\% are unstructured collapse.}
\label{fig:human_rollout_labels}
\end{figure}

We evaluate the official $\pi_{0.5}$ and GR00T-N1.7 checkpoints on LIBERO-PRO Spatial, Object, Goal, and Long (LIBERO-10). Every cell uses the instruction-delivery correction in \Cref{app:libero_pro_patch}. On both architectures, each nuisance cell exceeds each counterfactual cell within every suite, and Swap and Task are at least 42 points below ID (\Cref{fig:human_rollout_labels}, left, and \Cref{tab:libero_pro_diagnosis}). The pattern is therefore not unique to either of the evaluated architectures.

\subsection{Selecting a familiar behavior, then tracking its target}
\label{sec:human_rollout}
Success rates identify where a policy fails, but not what it does instead. We construct targeted perturbations (rules $R_{\mathrm{suite},n}$ in \Cref{tab:app_perturbation_rules}) and compare each rollout with the original behavior, the requested behavior, and other demonstrated tasks (\Cref{fig:human_rollout_labels}, right, and \Cref{tab:human_label_taxonomy}). Of 381 $\pi_{0.5}$ rollouts, 69\% follow a familiar behavior and 27\% follow or approximately follow the changed requirement. We call such rollouts \emph{retrieval-like}, a label for the observed match to a demonstrated behavior rather than a claim about the underlying mechanism. A second annotator independently labels 117 probes. Agreement on the retrieval-like distinction is $\kappa=0.78$ (\Cref{app:human_rollout}). Excluding the two rules where familiar behavior can remain valid, a 10~cm target shift ($R_{o,2}$) and reworded location clauses ($R_{s,2}$), leaves a retrieval-like share of 87\% across 291 probes.

The same 381 probes yield 73\% retrieval-like behavior and 14\% correct or near-correct following for GR00T-N1.7 (\Cref{tab:threeway_human}). On Goal, $\pi_{0.5}$ often executes another demonstrated task. On Long, it usually retains the source behavior, with occasional sub-trajectory hybrids (\Cref{fig:libero10_hybrid_example}). A nearest-trajectory check agrees with the human label on 88\% of usable Goal retrieval cases (\Cref{tab:goal_trajectory_nn}).

The Object probes separate target selection from tracking. Swapping object positions under the original instruction ($R_{o,1}$) often preserves the source-bound choice, whereas shifting the target by 10~cm ($R_{o,2}$) is followed visually. In all nine combined swapped-and-shifted probes, the policy tracks the displaced object occupying the source-bound slot, rather than the object named by the instruction. Thus these failures are neither fixed-coordinate replay nor a complete loss of visual feedback. The policy adjusts execution while retaining the wrong task-level choice.

\section{Task-Conditioned States Influence Action Selection}
\label{sec:mechanistic_evidence}
The other-task retrievals raise a question beyond sensitivity to language. When an instruction elicits the wrong familiar behavior, is that behavior encoded at the action interface? We test this with scene-matched readouts and interventions. \Cref{app:mechanistic_details} gives the full readout and intervention protocols, robustness checks, and controls.

\subsection{Readout and intervention on task-conditioned states}
\label{sec:kv_retrieval}
For a fixed set of 45 Goal rollouts labeled \texttt{same\_to\_other}, we construct ten candidate prefixes from the same source observation and each task's instruction. We rank candidates by cosine distance in the final-layer prefix key--value (KV) representation read by $\pi_{0.5}$'s action tokens. The executed task ranks first in $34/45$ cases and in the top three in $43/45$ (\Cref{fig:kv_executed_task}). A text-only TF-IDF comparison ranks it first in $22/45$. Holding the observation fixed makes this a readout of instruction-dependent task selection, with information beyond surface-word similarity.

\paragraph{Replacing the state redirects the arm.}
Readout is correlational, so we also intervene during closed-loop rollouts. At every policy call, we replace the prefix-KV cache with one computed from the target task's initial observation and instruction. Matched controls retain the source cache. The source scene, instruction, and policy are fixed. We enumerate all 50 ordered Object-suite task pairs whose target object is present in the source scene before evaluation, with four rollouts per condition. Under injection, $\pi_{0.5}$ ends nearer the target task's object in $109/200$ rollouts, versus $0/200$ controls. All 50 pairs shift more toward the target than their matched controls. Source-task completion falls from $200/200$ to zero (\Cref{tab:app_persistent_intervention}).

For GR00T-N1.7, we replace the backbone feature sequence read by the action head. Under the same protocol, $121/200$ rollouts end nearer the target object, versus none of the controls. Source-task completion again falls from $200/200$ to zero. Neither architecture completes the injected target task. The intervention changes the task-conditioned state, including its visual content, and removes access to a live representation. The measured effect is target-directed motion.

A complementary single-call intervention holds the observation fixed and substitutes each candidate instruction's final-layer cache for the 45 Goal queries. The patch-induced action displacement aligns more with the patched task's expert trajectory than with other tasks (cosine $0.44$ versus $0.18$ in \Cref{tab:kv_patch_directional}). The readouts link the executed task to the action interface, and interventions show that changing this state redirects behavior. Together with the probes in \Cref{sec:human_rollout}, they distinguish language sensitivity from selecting the action the task requires.

\section{Equivariant Counterfactual Training}
\label{sec:ect}
The diagnosis suggests a direct intervention. The instruction should be insufficient for choosing the action unless the policy also resolves the current scene. ECT data supply valid action alternatives under the same instruction, with scene differences that identify the appropriate action, and the ECT loss combines the imitation losses of corresponding alternatives in each update.

\begin{figure}[t]
    \centering
    \includegraphics[width=\linewidth]{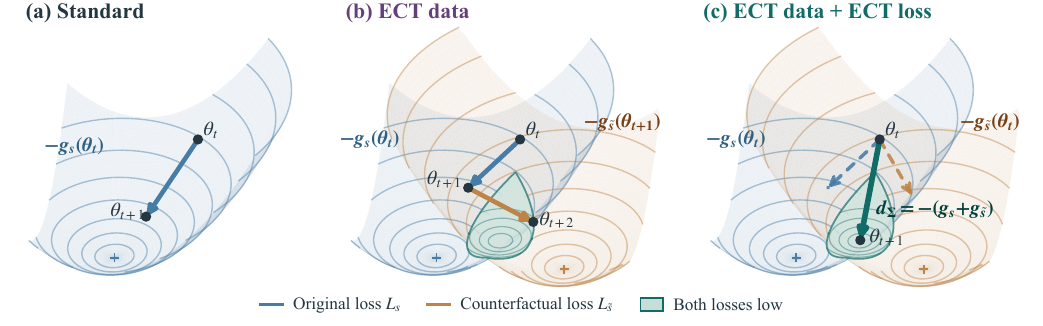}
    \vspace{-2.0em}
    \caption{\textbf{Intuition for ECT data and ECT loss.} Contours show the original and counterfactual losses. Shading marks a region where both are low.
    \textbf{(a)} Standard training uses only original demonstrations.
    \textbf{(b)} ECT data supply alternative supervision. The depicted independent-sampling sequence processes counterparts in separate updates. Independent batches can also contain both branches without matching counterparts.
    \textbf{(c)} The ECT loss evaluates matched counterparts at the same parameters and combines their gradients in one update. The paths show one illustrative geometry.}
    \label{fig:ect_pairing}
\end{figure}

\subsection{Method}
\label{sec:ect_method}

\begin{figure}[htbp]
\centering
\includegraphics[width=\linewidth]{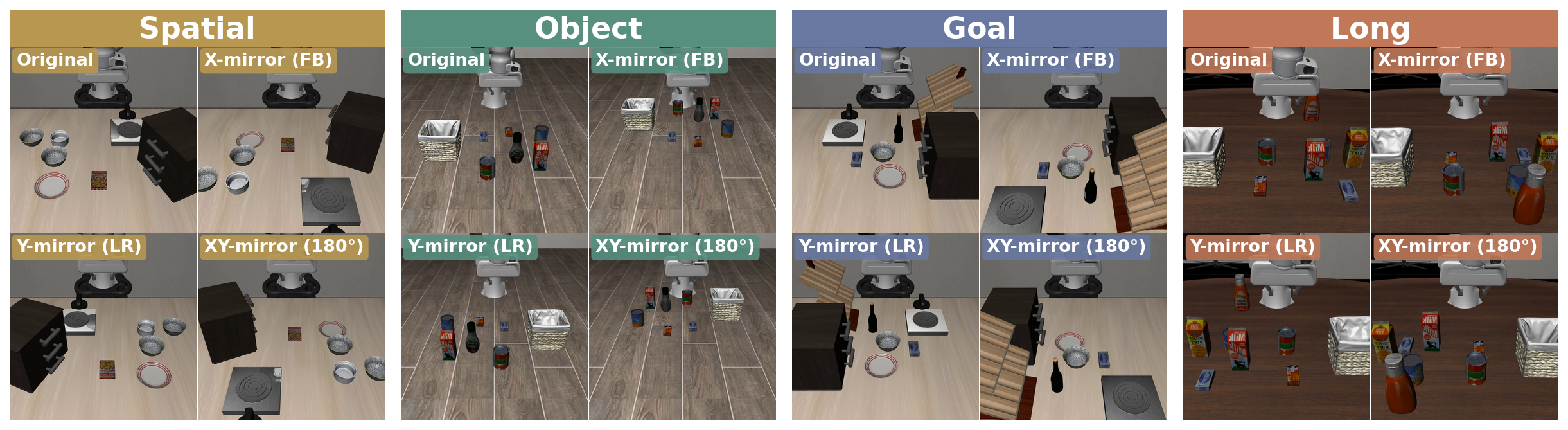}
\vspace{-2.0em}
\caption{\textbf{ECT scene constructions across the four LIBERO suites.} Each group shows an original scene and three mirrored versions under the same instruction. The training construction also uses shifts on Object and selected Long tasks. \Cref{tab:ect_transform_provenance} gives the transform families.}
\vspace{-0.5em}
\label{fig:ect_four_suites}
\end{figure}

\paragraph{Action-valid counterfactuals.}
We represent a pair as the five-tuple
\begin{equation}
e=(\ell,s,s',a,a'),\qquad (s,\ell,a)\ \text{and}\ (s',\ell,a')\ \text{are valid demonstrations}.
\label{eq:ect_tuple}
\end{equation}
The changed scene must require a different action under the same instruction, and the observations must distinguish the alternatives. In LIBERO, we construct the second demonstration by transforming and replaying the first (\Cref{fig:ect_four_suites}),
\begin{equation}
(s',\ell,a')=\mathcal{T}_{\mathrm{ECT}}(s,\ell,a),\qquad
s'=M_s(s),\quad a'=\mathrm{Replay}_{s'}\!\left(M_a(a)\right).
\label{eq:ect_two_branch}
\end{equation}
Here, $M_s$ changes the scene geometry and $M_a$ transforms the corresponding end-effector path. A replay controller executes the transformed path, records the resulting action, and retains successful demonstrations. This makes the second branch action-valid in the transformed scene rather than assuming equivariance of raw action commands. Appendix~\ref{app:ect_transform_details} details the construction.

\begin{figure}[t]
    \centering
    \includegraphics[width=0.64\linewidth]{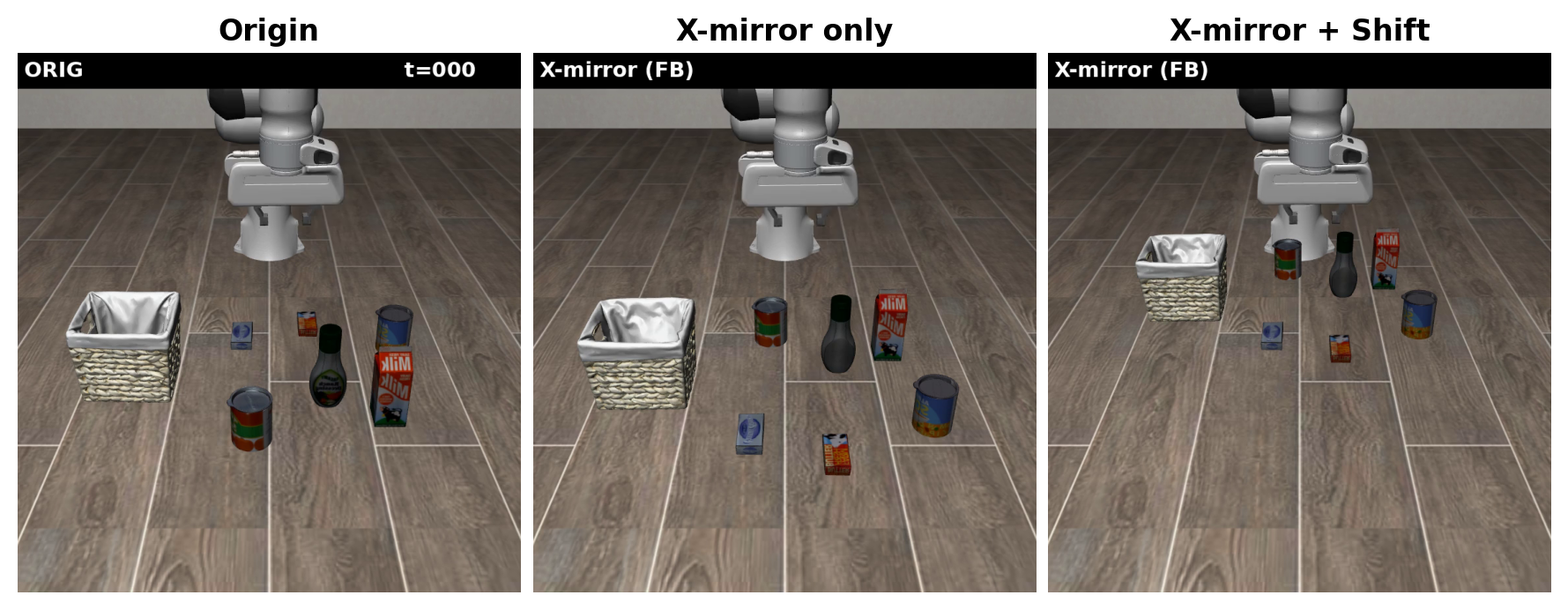}
    \vspace{-1.5em}
    \caption{\textbf{Useful counterfactuals must reveal the required action.} On Object, front-back mirroring changes the required action but leaves dominant landmarks in similar image regions. An added shift separates the branches visually. \Cref{tab:object_transform_diagnostics} reports performance by transformation.}
    \label{fig:ect_obj_shift_comp}
\end{figure}

\paragraph{What makes a useful counterfactual.}
Standard LIBERO supervision can make the instruction sufficient for selecting a familiar action even though the policy also observes the scene. ECT breaks this shortcut by holding $\ell$ fixed while making the correct action depend on the scene. Our first Object construction showed that this alone is not enough. Mirroring each scene front to back changes the required action but leaves the dominant landmarks in similar image regions, and ID success fell from 94\% to 24\% in per-suite diagnostic runs (\Cref{fig:ect_obj_shift_comp,tab:object_transform_diagnostics}). The binding account explains the failure. When $s$ and $s'$ are visually indistinguishable while the required actions differ, the pair removes the instruction-only solution without giving the policy a scene cue for selecting the correct branch. A useful pair therefore needs both $a\neq a'$ and observations that reveal which action is appropriate. ECT makes the instruction insufficient while keeping the scene sufficient for the choice.

A left-right mirror or added shift restores nominal competence by separating relevant landmarks in image space (\Cref{app:ect_transform_details}). Spatial and Goal use mirrors directly, while Object and selected Long tasks add shifts when needed. The front-back failure shows that action-changing diversity alone is insufficient. The scene must make the required action change identifiable.

The counterfactuals are built offline from training demonstrations, independently of LIBERO-PRO evaluation. \Cref{app:ect_no_leakage} compares their layouts with Swap states. The construction is one way to obtain the five-tuple in \Cref{eq:ect_tuple}. CALVIN supplies naturally occurring alternatives, and the UR5e study obtains them through physical demonstrations in different layouts. The paired procedure below applies to all three sources.

\paragraph{ECT loss.}
Let $\mathcal{E}$ be the collection of action-valid counterpart pairs. Standard training uses the original demonstrations, while the \emph{ECT data} baseline samples individual demonstrations independently from the enlarged training set. To optimize the \emph{ECT loss}, we instead sample matched pairs from $\mathcal{E}$ and sum the imitation losses of both counterparts in each update,
\begin{equation}
\mathcal{L}_{\mathrm{ECT}}(\theta)
=\mathbb{E}_{e\sim\mathcal{E}}\!\left[L_\theta(s,\ell,a)+L_\theta(s',\ell,a')\right],
\label{eq:ect_pair_loss}
\end{equation}
where $L_\theta$ is the standard per-example imitation loss, so the ECT loss adds no auxiliary objective. Each pair holds the instruction fixed while changing the scene and required action, and both counterparts contribute to the same parameter update. In our $\pi_{0.5}$ implementation, the two halves share the flow-matching noise sample and flow time. \Cref{tab:ect_decomposition} compares the ECT loss with independent sampling of the same demonstration pool. \Cref{fig:ect_pairing} illustrates the update construction, and \Cref{app:ect_transform_details} gives implementation details. \emph{ECT data + ECT loss} denotes the full method.

\subsection{Cross-architecture results}
\label{sec:ect_results}

\begin{table}[t]
\vspace{-1em}
\centering
\scriptsize
\setlength{\tabcolsep}{2pt}
\renewcommand{\arraystretch}{1.05}
\caption{\textbf{ECT counterfactual performance across architectures.} Success (\%). Frozen-LM controls match steps, batch size, and trainable parameters. Full-FT and GR00T use official checkpoints, not matched retrained controls. GR00T ECT models use the ECT data with the standard loss, and four-suite training changes scope (\Cref{app:groot_ect_results}). Parentheses give percentage-point gains over Standard (Frozen LM) or the official checkpoint (other blocks). ECT rows are shaded. Frozen-LM rows average three seeds, and other rows are single models.}
\label{tab:counterfactual_main}
\begin{tabular}{@{}l S[table-format=2.1]@{\hspace{2.5pt}}l S[table-format=2.1]@{\hspace{2.5pt}}l S[table-format=2.1]@{\hspace{2.5pt}}l S[table-format=2.1]@{\hspace{2.5pt}}l S[table-format=2.1]@{\hspace{2.5pt}}l S[table-format=2.1]@{\hspace{2.5pt}}l S[table-format=2.1]@{\hspace{2.5pt}}l S[table-format=2.1]@{\hspace{2.5pt}}l@{}}
\toprule
& \multicolumn{4}{c}{Spatial}
& \multicolumn{4}{c}{Object}
& \multicolumn{4}{c}{Goal}
& \multicolumn{4}{c}{Long} \\
\cmidrule(lr){2-5} \cmidrule(lr){6-9} \cmidrule(lr){10-13} \cmidrule(l){14-17}
Model & {Swap} & & {Task} &
      & {Swap} & & {Task} &
      & {Swap} & & {Task} &
      & {Swap} & & {Task} & \\
\midrule
\multicolumn{17}{@{}l}{\textit{$\pi_{0.5}$ Frozen LM (vision encoder and action expert trained)}} \\
Standard
& 52.7 & & 22.3 &
& 38.3 & & 23.0 &
& 41.3 & & 20.6 &
& 12.7 & & 19.5 & \\
\rowcolor{teal!7}
ECT data + ECT loss
& 74.4 & \bp{21.7} & 22.1 & \bn{0.2}
& 70.8 & \bp{32.5} & 31.3 & \bp{8.3}
& 55.5 & \bp{14.2} & 35.7 & \bp{15.1}
& 34.7 & \bp{22.0} & 22.5 & \bp{3.0} \\
\midrule
\multicolumn{17}{@{}l}{\textit{$\pi_{0.5}$ Full FT (all parameters trained)}} \\
Official $\pi_{0.5}$
& 46.6 & & 53.0 &
& 18.2 & & 11.0 &
& 34.2 & & 21.2 &
& 9.4 & & 17.0 & \\
\rowcolor{teal!7}
ECT data + ECT loss
& 72.8 & \bp{26.2} & 53.2 & \bp{0.2}
& 40.8 & \bp{22.6} & 28.4 & \bp{17.4}
& 35.8 & \bp{1.6} & 26.6 & \bp{5.4}
& 26.6 & \bp{17.2} & 26.6 & \bp{9.6} \\
\midrule
\multicolumn{17}{@{}l}{\textit{GR00T-N1.7}} \\
Official (per suite)
& 1.2 & & 51.4 &
& 0.0 & & 9.0 &
& 2.2 & & 10.0 &
& 0.2 & & 10.2 & \\
\rowcolor{teal!7}
ECT data, per suite
& 17.8 & \bp{16.6} & 53.0 & \bp{1.6}
& 0.0 & \bz{0.0} & 10.0 & \bp{1.0}
& 16.2 & \bp{14.0} & 10.0 & \bz{0.0}
& 9.6 & \bp{9.4} & 10.4 & \bp{0.2} \\
\rowcolor{teal!7}
ECT data, four suites
& 38.2 & \bp{37.0} & 64.0 & \bp{12.6}
& 7.0 & \bp{7.0} & 10.2 & \bp{1.2}
& 17.4 & \bp{15.2} & 10.6 & \bp{0.6}
& 16.2 & \bp{16.0} & 13.2 & \bp{3.0} \\
\bottomrule
\end{tabular}
\end{table}

In the controlled $\pi_{0.5}$ comparison, ECT raises mean Swap success from 36\% to 59\% and mean Task success from 21\% to 28\% (\Cref{tab:counterfactual_main}). Both models use the Frozen-LM recipe, training the vision encoder and action expert with a fixed language model, matched steps, and matched batch size. \Cref{tab:counterfactual_main} also reports full-fine-tuning and GR00T comparisons against official checkpoints, with their comparison conditions specified in the caption and \Cref{app:groot_ect_results}.

The benefits extend beyond the spatial changes directly constructed during training. For example, Frozen-LM $\pi_{0.5}$ improves Goal Task by 15 points, while Full FT improves Object Task by 17 points. On the 381 behavioral probes, the ECT model follows or approximately follows the changed requirement in 48\% of rollouts, versus 27\% for the official $\pi_{0.5}$, and retrieval-like behavior falls from 69\% to 47\% (\Cref{tab:threeway_human}). The observed gain includes a shift away from familiar but incorrect behaviors. Here, the official Full-FT model is compared with Frozen-LM ECT, so data, paired training, and the fine-tuning recipe change together.

Adaptation also depends on the requested distinction. For new Spatial destinations, erroneous completion of the original placement falls from $29/30$ to $3/30$, while both models still pick the bowl after an object-name change (\Cref{app:probe_retrieval_after_ect}). The training data vary where destinations occur but keep the bowl in every Spatial instruction. This selectivity is consistent with the supervision account. ECT improves distinctions exposed by the added alternatives, but does not create distinctions that remain absent from the training data. This suggests collecting alternatives that expose the particular object, goal, or relation a policy must distinguish.

\paragraph{ECT data and ECT loss are complementary.}
\label{sec:ect_decomposition}
ECT data create missing scene-dependent alternatives, while the ECT loss makes their correspondence explicit during training. Although the data component accounts for most of the LIBERO Swap gain, the ECT loss adds a further 2 to 5 points on Spatial, Goal, and Long (\Cref{tab:ect_decomposition}). On CALVIN, where counterparts already exist in the original data, the ECT loss alone raises five-task completion by 18 points (\Cref{tab:beyond_libero}).

\begin{table}[t]
\centering
\scriptsize
\setlength{\tabcolsep}{3pt}
\renewcommand{\arraystretch}{1.0}
\caption{\textbf{ECT data and ECT loss make distinct contributions.}
Swap success (\%). All rows use Frozen-LM training with matched steps and batch size (64 examples per update). Each row averages three training seeds (\Cref{app:ect_decomposition}). Pairing draws its pairs from the same pool as ECT data. Full five-cell results are in \Cref{tab:app_pi05_unified_ect}. ECT rows are shaded, with the best result per column in bold.}
\label{tab:ect_decomposition}
\begin{tabular}{@{}lcccc@{}}
\toprule
 & Spatial & Object & Goal & Long \\
\midrule
Standard & 52.7 & 38.3 & 41.3 & 12.7 \\
\rowcolor{teal!7}
ECT data & 72.1 & \textbf{71.1} & 51.8 & 29.5 \\
\rowcolor{teal!7}
ECT data + ECT loss & \textbf{74.4} & 70.8 & \textbf{55.5} & \textbf{34.7} \\
\midrule
$\Delta_{\text{data}}$ (ECT data $-$ Standard) & $+19.4$ & $+32.8$ & $+10.5$ & $+16.8$ \\
$\Delta_{\text{loss}}$ (ECT data + ECT loss $-$ ECT data) & $+2.3$ & $-0.3$ & $+3.7$ & $+5.2$ \\
\bottomrule
\end{tabular}

\par\vspace{\floatsep}
\centering
\scriptsize
\setlength{\tabcolsep}{3pt}
\renewcommand{\arraystretch}{1.0}
\vspace{-1em}
\caption{\textbf{ECT beyond LIBERO.} The top block reports real-robot UR5e results with 400 demonstrations per main model. The ECT variants differ in paired training. $^\dagger$Swap uses companion models excluding the overlapping training layout (\Cref{app:ur5}). The bottom block reports $\pi_{0.5}$ on CALVIN ABC$\to$D (\Cref{app:calvin}) with one training seed per row and 1{,}000 evaluation sequences. LH-$k$ is the rate of completing $k$ tasks in a row. ECT rows are shaded, with the best result per column in bold.}
\label{tab:beyond_libero}
\begin{tabular}{@{}l cccccc@{}}
\toprule
\textit{UR5e (real robot)} & & \multicolumn{2}{c}{nuisance} & \multicolumn{3}{c}{counterfactual} \\
\cmidrule(lr){3-4}\cmidrule(l){5-7}
 & ID & Paraphrase & Attribute & Unseen pos. & Unseen obj.\,+\,pos. & Swap \\
\midrule
Standard & \textbf{40/40} & 19/20 & 9/10 & 3/40 & 0/10 & 4/20 \\
\rowcolor{teal!7}
ECT data & 37/40 & \textbf{20/20} & 9/10 & 30/40 & \textbf{8/10} & 12/14$^\dagger$ \\
\rowcolor{teal!7}
ECT data + ECT loss & 37/40 & \textbf{20/20} & \textbf{10/10} & \textbf{35/40} & \textbf{8/10} & \textbf{20/20}$^\dagger$ \\
\midrule
\multicolumn{7}{@{}l}{\textit{CALVIN ABC$\to$D}} \\
 & LH-1 & LH-3 & LH-5 & Avg.\ len. & & \\
\midrule
Standard & 89.2 & 72.2 & 57.6 & 3.63 & & \\
\rowcolor{teal!7}
Original data + ECT loss & 95.2 & \textbf{85.5} & \textbf{75.6} & \textbf{4.28} & & \\
\rowcolor{teal!7}
ECT data & 94.2 & 79.2 & 65.7 & 3.98 & & \\
\rowcolor{teal!7}
ECT data + ECT loss & \textbf{96.5} & 84.3 & 70.6 & 4.21 & & \\
\bottomrule
\end{tabular}

\end{table}

\subsection{Beyond LIBERO, and beyond more data}
\label{sec:beyond}
\paragraph{Physical demonstrations under a fixed budget.}
The UR5e study allocates 400 demonstrations either to repetitions in one layout or to four layouts that require different reaches under the same instruction. \Cref{app:ur5} details the robot setup, with training layouts and evaluation conditions illustrated in \Cref{fig:ur5_settings}. On unseen positions, Standard succeeds in $3/40$ trials, ECT data in $30/40$, and the paired method in $35/40$, while ID success remains high (\Cref{tab:beyond_libero}). These positions occur in none of the training layouts. Prior work shows the value of environment and object diversity over repeated demonstrations \citep{ICLR2025_88b7b2c8}. Here, under the same fixed budget, we specifically vary scenes so that the same instruction requires different actions, with an additional benefit from paired training.

\paragraph{ECT loss on existing demonstrations.}
CALVIN ABC$\to$D tests the ECT loss without generating data. Its original demonstrations already contain the same instructions in different scenes that require different actions. The ECT loss alone raises $\pi_{0.5}$'s five-task completion from 58\% to 76\%, without constructing additional demonstrations. Constructed data also improve over Standard, and the ECT loss adds to that result. Natural pairs perform best under the reported budget. \Cref{app:calvin} describes the construction, evaluation, and data-coverage budgets for these single-seed comparisons.

\paragraph{Repetition does not substitute for the missing alternatives.}
At fixed 30k steps, increasing the original LIBERO demonstrations from 25\% to 100\% leaves every Swap and Task cell at least 34 points below ID (\Cref{tab:sweep}). Standard RL post-training also preserves much of ECT's advantage (\Cref{app:pirl}). Together with CALVIN and the fixed-budget robot study, these results separate scene-dependent action alternatives from demonstration count alone.

\section{Related Work and Conclusion}
\label{sec:related_work_discussion}
\label{sec:conclusion}
\paragraph{Diagnosing VLA generalization.}
Recent work studies language under-use, modality imbalance, and visual changes during VLA fine-tuning \citep{lian2026bayesianvla,fei2025libero,xu2025seeing,darabi2026progal,kachaev2025don}. \citet{fang2026vision} report familiar-task execution under counterfactual instructions, attribute it to vision shortcuts, and introduce Counterfactual Action Guidance (CAG). Complementing this account, our Goal probes and scene-matched interventions show that instructions can select the wrong familiar behavior, while Object probes show retained local tracking. CAG strengthens language conditioning at inference; ECT supplies scene-dependent action alternatives and trains counterparts together.

\paragraph{Shortcut learning and counterfactual supervision.}
Instruction-action binding connects to shortcut learning and underspecification, where training performance does not determine which predictive dependencies a model uses \citep{geirhos2020shortcut,d2022underspecification}. Counterfactual augmentation exposes missing distinctions in dynamics, visual classification, and language tasks \citep{pitis2020counterfactual,chang2021towards,Kaushik2020Learning}. Beyond adding examples, \citet{teney2020learning} use differences between counterfactual inputs to supervise gradient orientation through an auxiliary objective. ECT uses counterpart correspondence in batch construction while retaining the standard imitation loss for each example. Same-pool comparisons test whether the ECT loss improves generalization.

\paragraph{Constructing robot demonstrations.}
CAST \citep{glossop2025cast} changes instructions and actions for a fixed observation, whereas ECT fixes the instruction and varies the scene and required action. RoCoDA \citep{ameperosa2025rocoda} and MimicGen \citep{mandlekar2023mimicgen} also construct demonstrations in new scene configurations. ECT shares these geometric mechanisms but uses them to address the supervision gap identified by instruction-action binding. Successful replay ensures action-valid counterparts, and observable scene differences make the required alternatives distinguishable. The ECT loss preserves their correspondence during training. Unlike equivariant robot policies \citep{yang2024equibot,wang2024equivariant}, ECT keeps the policy architecture unchanged. CALVIN further applies the ECT loss to counterparts already present in the data.

\paragraph{Conclusion.}
VLA policies can respond to language and retain local visual tracking while selecting a familiar behavior that no longer satisfies the task. Our behavioral analyses and internal-state interventions connect this gap to instruction-action binding, which narrow conditional action support in the demonstrations permits. ECT addresses this supervision gap through ECT data and the ECT loss. ECT data provide valid demonstrations in distinguishable scenes requiring different actions under the same instruction, and the ECT loss trains corresponding alternatives together. The LIBERO, CALVIN, and UR5e experiments show the value of constructing missing alternatives, exploiting existing counterparts, and collecting scene-dependent demonstrations. Counterfactual evaluation is therefore essential for distinguishing genuine task-level grounding from policies that remain robust while selecting familiar but incorrect behaviors.

\clearpage
\subsection*{Reproducibility statement}
Training recipes and trainable parameter groups appear in \Cref{tab:recipes}.
\Cref{app:ect_details,app:ect_libero_results,app:ect_beyond_libero} document the configurations, data versions, and ECT construction, including the UR5e and CALVIN protocols in \Cref{app:ur5,app:calvin}.
The corrected LIBERO-PRO loader and its validation are in \Cref{app:libero_pro_patch}.
The annotation protocol is in \Cref{app:human_rollout}, and seeds and uncertainty are reported in \Cref{app:uncertainty}.
\subsection*{AI use statement}
We used generative AI tools to provide feedback on the conceptual framework, mathematical formulation, and experimental methodology, to assist with interpreting results, and to help implement standard components (data loaders, evaluation scripts, and plotting) and run and log experiments. We also used these tools to draft and edit text, organize the manuscript, and develop or revise scientific figures. The authors verified the experimental designs, data, results, and claims. Every reported number is traced to a result file and a checkpoint, and all AI-assisted text was reviewed for accuracy. We take responsibility for the final content of this work, including text, claims, code, and figures produced with the aid of generative AI.

\bibliography{references}
\bibliographystyle{iclr2027_conference}

\newpage
\appendix
\crefalias{section}{appendix}\crefalias{subsection}{appendix}\crefalias{paragraph}{appendix}   
\section*{Notation Table}
\begin{table*}[htbp]
\centering
\scriptsize
\setlength{\tabcolsep}{3.5pt}
\renewcommand{\arraystretch}{1.0}
\caption{\textbf{Frequently used notation in the main paper and appendix.}}
\begin{tabular}{@{}p{0.22\linewidth}p{0.73\linewidth}@{}}
\toprule
Symbol & Meaning \\
\midrule
\multicolumn{2}{@{}l}{\textit{Instruction-action binding framework}} \\
$\mathcal{D}$ & Demonstration dataset. \\
$(s_i,\ell_i,a_i)$ & Observation, instruction, and expert trajectory from task $i$. \\
$T$ & Task identity. \\ $H(T\mid x)$ & Conditional entropy: remaining uncertainty about task identity $T$ after observing cue $x$, such as instruction $\ell$ or observation $s$. \\
$a^\star(s,\ell)$ & Task-level action requirement, abstracting from incidental execution variation. \\
$\pi(a\mid s,\ell)$ & Language-conditioned policy. \\
$\pi_{\mathrm{ground}}$, $\pi_{\mathrm{lookup}}$ & Idealized grounded policy and instruction-keyed lookup policy. \\
$r(\ell)$, $g(s,r(\ell))$ & Linguistic referent of instruction $\ell$ and its grounded referent in scene $s$. \\
$k(\ell)$ & Instruction-keyed task selection in the conceptual lookup model. \\
$f_\ell(\ell)$, $d_\ell(\cdot,\cdot)$ & Instruction representation and distance used in the conceptual lookup model. \\
$\Delta_k,\Delta_a$ & Changes in the selected instruction key and the task-level action requirement. \\
\midrule
\multicolumn{2}{@{}l}{\textit{Prefix-KV retrieval and patching}} \\
$K^{(L)}(s,\ell), V^{(L)}(s,\ell)$ & Final-layer prefix key and value tensors for scene-instruction input $(s,\ell)$. \\
$\phi_{\mathrm{KV}}(s,\ell)$, $d_{\mathrm{KV}}(x,y)$ & Flattened normalized prefix-KV representation and cosine distance. \\
$q=(s_{\mathrm{src}},\ell_{\mathrm{pert}})$ & Perturbed query input for prefix-KV retrieval. \\
$x_c=(s_{\mathrm{src}},\ell_c)$ & Candidate prefix input using the same source scene and candidate instruction $\ell_c$. \\
$c_{\mathrm{exe}}$, $\mathrm{rank}_{\mathrm{exe}}(q)$ & Human-labeled executed task and its nearest-neighbor rank in prefix-KV space. \\
$A_{\mathrm{nat}}$, $A_{\mathrm{patch}}$, $A_T$ & Natural decoded action chunk, patched decoded action chunk, and canonical expert chunk for target task $T$. \\
$\Delta_{\mathrm{L2}}(T)$ & L2 improvement from patching toward target task $T$. \\
\midrule
\multicolumn{2}{@{}l}{\textit{Equivariant Counterfactual Training and objectives}} \\
$e=(\ell,s,s',a,a')$, $\mathcal{E}$ & Action-valid counterpart pair and the collection sampled by paired training. \\
$\mathcal{T}_{\mathrm{ECT}}$ & ECT operator mapping a demonstration to an action-valid counterfactual. \\
$M_s$, $M_a$ & Scene transformation and the matching transformation of the end-effector path. \\
$\mathrm{Replay}_{\tilde{s}}$ & Tracks a transformed path in scene $\tilde{s}$ and returns the recorded action. \\
$(\tilde{s},\ell,\tilde{a})=(s',\ell,a')$ & Counterpart demonstration. Tildes denote the constructed branch in this appendix. \\
$L_\theta(s,\ell,a)$, $\mathcal{L}_{\mathrm{ECT}}(\theta)$ & Per-example action-imitation loss and the ECT loss. \\
$a_0$, $a_t$, $t$ & Noise sample, interpolated action, and flow time. \\
$v_\theta(a_t,t\mid s,\ell)$ & Flow-matching velocity field conditioned on scene and instruction. \\
$\mathcal{L}_{\mathrm{FM}}$ & Flow-matching imitation loss. \\
\midrule
\multicolumn{2}{@{}l}{\textit{Human-label provenance}} \\
$R_{g,n},R_{10,n}$, $R_{o,n},R_{s,n}$ & Suite-local rollout-construction rule identifiers for Goal, Long, Object, and Spatial. \\
\bottomrule
\end{tabular}
\end{table*}
\renewcommand{\arraystretch}{1.0}

\section*{Supplementary Overview}
The appendix provides the following details.
\begin{itemize}[nosep,leftmargin=*]
\item \hyperref[app:experimental_details]{\textbf{\S\ref*{app:experimental_details}}}\quad Experimental setup and success-rate conventions.
\item \hyperref[app:libero_pro_patch]{\textbf{\S\ref*{app:libero_pro_patch}}}\quad Corrected LIBERO-PRO evaluation and validation.
\item \hyperref[app:objective_interpretation]{\textbf{\S\ref*{app:objective_interpretation}}}\quad Instruction-keyed solutions under the flow-matching objective.
\item \hyperref[app:human_rollout]{\textbf{\S\ref*{app:human_rollout}}}\quad Annotation protocol, reliability, behavior counts, stress tests, and post-ECT probes.
\item \hyperref[app:mechanistic_details]{\textbf{\S\ref*{app:mechanistic_details}}}\quad Prefix readouts, intervention protocols, and real-robot offline analysis.
\item \hyperref[app:ect_details]{\textbf{\S\ref*{app:ect_details}}}\quad ECT data construction, action validity, and separation from the evaluation layouts.
\item \hyperref[app:ect_libero_results]{\textbf{\S\ref*{app:ect_libero_results}}}\quad Complete LIBERO-PRO results, decomposition, data-scale sweep, and RL post-training.
\item \hyperref[app:ect_beyond_libero]{\textbf{\S\ref*{app:ect_beyond_libero}}}\quad ECT beyond LIBERO: the physical UR5e study and CALVIN ABC$\to$D.
\item \hyperref[app:uncertainty]{\textbf{\S\ref*{app:uncertainty}}}\quad Statistical uncertainty and seed-level stability.
\item \hyperref[app:other_accounts]{\textbf{\S\ref*{app:other_accounts}}}\quad Relation to other accounts of VLA failure.
\item \hyperref[app:limitations]{\textbf{\S\ref*{app:limitations}}}\quad Limitations.
\end{itemize}

\section{Experimental Details}
\label{app:experimental_details}

We evaluate two VLA families, $\pi_{0.5}$ and GR00T-N1.7. By default, $\pi_{0.5}$ is the official checkpoint, fully fine-tuned on all four LIBERO suites. GR00T-N1.7 provides cross-architecture validation of the LIBERO-PRO failure pattern, the human rollout study, the persistent intervention, and ECT. 

The four suites are LIBERO-Spatial, LIBERO-Object, LIBERO-Goal, and LIBERO-Long (also called LIBERO-10), written Spatial, Object, Goal, and Long. LIBERO-PRO adds controlled perturbations in five cells, of which we use Semantic, object perturbation, Swap, and Task (\Cref{app:libero_pro_patch}). Semantic paraphrases and object perturbations usually leave the required trajectory unchanged, whereas Swap and Task require recomputing the action from the current scene and instruction.

Success is the simulator predicate at episode termination, not a transient first reach or contact. Full-suite evaluations run 50 trials on each of the 10 tasks per suite ($N=500$ per cell). Human rollout labels classify behavior and do not replace simulator success (\Cref{tab:app_eval_components}). Implementation otherwise follows each model family's original pipeline.

\begin{table}[htbp]
\centering
\scriptsize
\setlength{\tabcolsep}{5pt}
\caption{\textbf{Evaluation components.} All LIBERO success-rate tables use terminal-state success. Human labels and mechanistic diagnostics are separate analyses.}
\label{tab:app_eval_components}
\begin{tabular}{lll}
\toprule
Experiment & Evaluation unit & Measurement \\
\midrule
LIBERO-PRO diagnosis & Perturbation rollout & Terminal-state simulator predicate \\
Human rollout labels & Rollout video & Behavioral category \\
Prefix-KV retrieval & \texttt{same\_to\_other} rollout & Nearest-neighbor executed-task recovery \\
Persistent intervention & Rollout & End-effector proximity and simulator predicate \\
\bottomrule
\end{tabular}
\end{table}

\begin{table}[htbp]
\centering
\scriptsize
\renewcommand{\arraystretch}{0.95}
\caption{\textbf{LIBERO-PRO success of the official checkpoints.} $\pi_{0.5}$ uses one four-suite checkpoint and GR00T-N1.7 uses per-suite checkpoints. We evaluate $N=500$ episodes per cell. \Cref{fig:human_rollout_labels} (left) shows the four-suite means. Sem.\ and Task deliver the perturbed instruction to the policy (\Cref{app:libero_pro_patch}).}
\label{tab:libero_pro_diagnosis}
\begin{tabular*}{\linewidth}{@{\extracolsep{\fill}}l ccccc ccccc@{}}
\toprule
 & \multicolumn{5}{c}{$\pi_{0.5}$} & \multicolumn{5}{c}{GR00T-N1.7} \\
\cmidrule(lr){2-6}\cmidrule(l){7-11}
Suite & ID & Sem. & Obj. & Swap & Task & ID & Sem. & Obj. & Swap & Task \\
\midrule
Spatial & 98.4 & 98.0 & 98.0 & 46.6 & 53.0 & 93.6 & 88.8 & 93.0 & 1.2 & 51.4 \\
Object  & 98.6 & 99.0 & 94.4 & 18.2 & 11.0 & 95.6 & 97.6 & 86.0 & 0.0 & 9.0 \\
Goal    & 96.4 & 94.6 & 88.8 & 34.2 & 21.2 & 94.2 & 93.8 & 76.0 & 2.2 & 10.0 \\
Long      & 91.8 & 90.8 & 66.8 & 9.4 & 17.0 & 89.0 & 88.4 & 54.6 & 0.2 & 10.2 \\
\bottomrule
\end{tabular*}
\end{table}

\section{LIBERO-PRO Evaluation Protocol}
\label{app:libero_pro_patch}

For its Semantic and Task cells, LIBERO-PRO writes the perturbed instruction into the \texttt{(:language ...)} block of each regenerated BDDL file, but its released evaluation code reads the instruction from the file name, so the policy receives the original instruction while the success predicate scores the perturbed task. The maintainers have acknowledged this issue in the official LIBERO-PRO repository\footnote{\url{https://github.com/Zxy-MLlab/LIBERO-PRO/issues/14}}, and, following the fix recommended there, we read these two cells' instructions from the \texttt{:language} block. The ID, object-perturbation, and Swap cells do not perturb the instruction and keep the filename-derived wording used in the demonstrations. Simulator goals and success predicates are unchanged, and every number in this paper uses this protocol.

A scripted oracle parses the target object from the delivered instruction and executes a pick-and-place with ground-truth poses, scoring $90/100$ on the Object Task cell, where the official $\pi_{0.5}$ scores $11.0\%$. This checks instruction delivery and task executability rather than measuring learned grounding. LIBERO-PRO's fifth cell, Environment, perturbs scene assets that are not included in the public release\footnote{\url{https://github.com/Zxy-MLlab/LIBERO-PRO/issues/9}}, so no Environment numbers appear in this paper.

\section{Objective-Level Interpretation}
\label{app:objective_interpretation}
\label{app:flow_matching_lookup}
We relate the binding framework of \Cref{sec:framework} to the flow-matching objective used by both evaluated model families. Under the condition below, the instruction key can predict the demonstrated action without resolving how the required action changes with the scene.

A flow-matching policy \citep{lipman2022flow, intelligence2025pi_, bjorck2025gr00t} uses observation $s$, instruction $\ell$, expert trajectory $a$, and noise $a_0$. For $t\in[0,1]$, its interpolated action is
\begin{equation}
a_t=(1-t)a_0+t a ,
\label{eq:app_flow_interpolation}
\end{equation}
and the flow-matching loss is
\begin{equation}
\mathcal{L}_{\mathrm{FM}}(\theta)
=
\mathbb{E}_{(s,\ell,a),t,a_0}
\left[
\left\|
v_\theta(a_t,t\mid s,\ell) - (a-a_0)
\right\|_2^2
\right].
\label{eq:app_flow_matching_loss}
\end{equation}
Under squared loss, the population minimizer of the velocity field $v_\theta$ is the conditional expectation
\begin{equation}
v^*(a_t,t\mid s,\ell)
=
\mathbb{E}\left[
a-a_0
\mid
a_t,t,s,\ell
\right].
\label{eq:app_flow_population_minimizer}
\end{equation}
If $k(\ell)$ nearly determines the expert trajectory on the training distribution, then
\begin{equation}
\mathbb{E}\left[
a-a_0
\mid
a_t,t,s,\ell
\right]
\approx
\mathbb{E}\left[
a-a_0
\mid
a_t,t,k(\ell)
\right],
\label{eq:app_flow_key_sufficiency}
\end{equation}
so an instruction-conditioned velocity field can achieve low imitation loss,
\begin{equation}
v_\theta(a_t,t\mid s,\ell)
\approx
v_\theta(a_t,t\mid k(\ell)).
\label{eq:app_flow_lookup_solution}
\end{equation}
The scene can still shape token representations and local execution details, but when instruction identity predicts the demonstrated trajectory on the training distribution, the objective alone does not force counterfactual scene-conditioned recomposition.

\section{Human Rollout Evaluation}
\label{app:human_rollout}

This appendix gives the annotation protocol and detailed label results behind \Cref{sec:human_rollout}.

\subsection{Annotation protocol}

\begin{figure}[htbp]
\centering
\includegraphics[width=\linewidth]{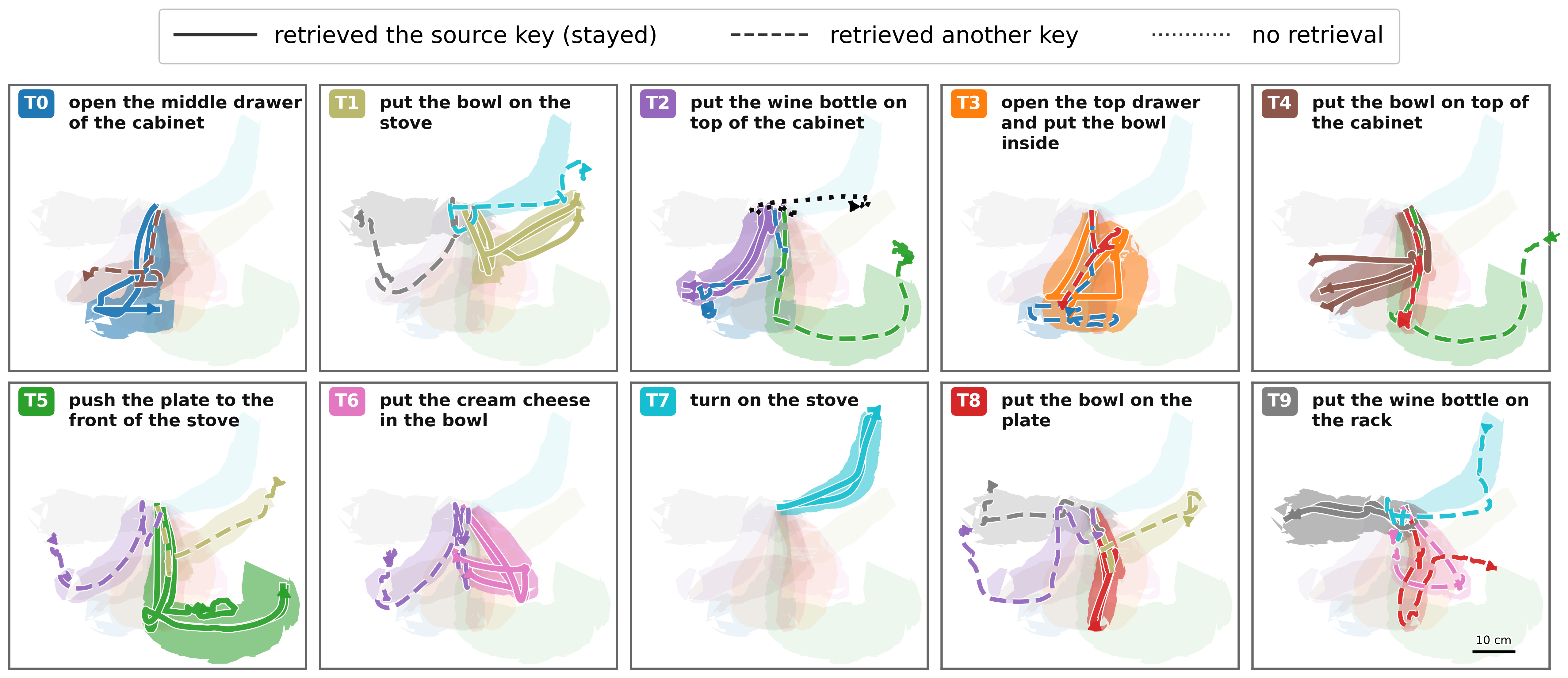}
\caption{\textbf{Goal rollouts can follow familiar training trajectories.} Medoid rollouts for the Goal perturbation rules are overlaid on the 80th-percentile envelopes of the corresponding tasks' 50 training demonstrations. The medoids are selected from 117 rollouts under $R_{g,1}$ to $R_{g,4}$. Task 3 groups destinations for readability. The comparisons illustrate whole-task matches.}
\label{fig:goal_trajectory_distribution}
\end{figure}

The annotator compares each perturbed rollout with the source behavior, the behavior required by the modified instruction or scene, and other demonstrated tasks. Each rollout receives one fine-grained label (\Cref{tab:human_label_taxonomy}). The main analysis groups these labels into retrieval-like behavior, correct or near-correct following, and collapse. In \Cref{fig:goal_trajectory_distribution}, each Goal task shows the medoid rollout of each rule, drawn from the 117 rollouts of $R_{g,1}$ to $R_{g,4}$, against the 80th-percentile envelope of its 50 training demonstrations. Task 3 groups destinations for readability.

\begin{table}[H]
\centering
\scriptsize
\setlength{\tabcolsep}{3.2pt}
\caption{\textbf{Human rollout annotation taxonomy.}
Fine labels map to the main paper's coarse groups.}
\label{tab:human_label_taxonomy}
\begin{tabular}{lll}
\toprule
Fine-grained label & Coarse group & Definition \\
\midrule
\texttt{same\_to\_origin} & Source retrieval & Clearly executes the original source behavior \\
\texttt{close\_to\_origin} & Source retrieval & Approximately follows the source behavior \\
\texttt{same\_to\_other} & Other-task retrieval & Matches another familiar demonstrated task \\
\texttt{same\_to\_change} & Correct or near-correct & Executes the modified instruction correctly \\
\texttt{close\_to\_change} & Correct or near-correct & Approximately follows the modified instruction \\
\texttt{mix} & Hybrid retrieval & Combines source, changed, or neighboring sub-trajectories \\
\texttt{robot\_collapse} & Collapse & Fails without a coherent retrieval-like trajectory \\
\bottomrule
\end{tabular}
\end{table}

\paragraph{Annotation reliability.}
A primary annotator labeled all 381 rollouts for the main analysis. A second annotator independently labeled the first episode of each rollout construction, giving 117 doubly labeled rollouts as a reliability check. \Cref{tab:annotation_reliability} reports agreement for the seven fine labels, the five coarse groups of \Cref{fig:human_rollout_labels}, and the retrieval-like versus correct binary on which the main claim rests. Agreement rises as labels coarsen and is highest on the binary. Per-suite coarse-label $\kappa$, computed on small subsamples and therefore only indicative, ranges from $0.65$ on Goal to $0.75$ on Spatial.

\paragraph{GR00T-N1.7 and ECT labels.}
The same 381 probes were labeled with the same rubric for the official GR00T-N1.7 and for $\pi_{0.5}$ trained with ECT (\Cref{tab:threeway_human}). For both official policies, whose action pathways share no component, following a familiar trajectory family is the dominant outcome. GR00T-N1.7 also collapses more often than $\pi_{0.5}$, mostly on Goal.

\begin{table}[t]
\centering
\scriptsize
\setlength{\tabcolsep}{5pt}
\caption{\textbf{Inter-annotator reliability.} 117 of the 381 probes were labeled by both annotators. Brackets give 95\% episode-bootstrap confidence intervals.}
\label{tab:annotation_reliability}
\begin{tabular}{lcccc}
\toprule
Granularity & $N$ & Agreement & Cohen's $\kappa$ & Macro F1 \\
\midrule
Fine labels
& 117 & 73.5 [65.0, 81.2] & 0.611 [0.504, 0.712] & 45.9 [35.5, 54.0] \\
Coarse 5-class
& 117 & 83.8 [76.9, 90.6] & 0.731 [0.617, 0.833] & 58.5 [47.0, 67.6] \\
Retrieval-like binary
& 117 & 90.6 [84.6, 95.7] & 0.778 [0.639, 0.894] & 88.9 [81.9, 94.7] \\
\bottomrule
\end{tabular}
\end{table}

\subsection{Aggregate and rule-level outcomes}
\label{app:human_aggregate_rule_results}

\Cref{tab:app_human_rule_counts} reports the aggregate counts. Of 381 probes, 69.3\% show source-task, other-task, or hybrid retrieval, and 26.5\% show correct or near-correct following.

\begin{table}[htbp]
\centering
\scriptsize
\setlength{\tabcolsep}{3.5pt}
\caption{\textbf{Human-labeled outcomes of the official $\pi_{0.5}$ by rule and suite.} Goal subrules are aggregated into their parent rules. $R_{s,2}$ pools two variants of the reworded location.}
\label{tab:app_human_rule_counts}
\begin{tabular}{llrrrrrr}
\toprule
Suite & Rule & $N$ & Source & Other-task & Hybrid & Correct & Collapse \\
\midrule
Goal & $R_{g,1}$ & 33 & 27 & 2 & 0 & 0 & 4 \\
 & $R_{g,2}$ & 33 & 12 & 12 & 0 & 9 & 0 \\
 & $R_{g,3}$ & 21 & 6 & 9 & 0 & 6 & 0 \\
 & $R_{g,4}$ & 30 & 7 & 21 & 0 & 0 & 2 \\
 & $R_{g,5}$ & 3 & 3 & 0 & 0 & 0 & 0 \\
 & \emph{total} & \textbf{120} & \textbf{55} & \textbf{44} & \textbf{0} & \textbf{15} & \textbf{6} \\
\midrule
Long & $R_{10,1}$ & 30 & 24 & 0 & 0 & 5 & 1 \\
 & $R_{10,2}$ & 30 & 29 & 0 & 1 & 0 & 0 \\
 & $R_{10,3}$ & 12 & 3 & 2 & 4 & 2 & 1 \\
 & \emph{total} & \textbf{72} & \textbf{56} & \textbf{2} & \textbf{5} & \textbf{7} & \textbf{2} \\
\midrule
Object & $R_{o,1}$ & 30 & 25 & 2 & 0 & 3 & 0 \\
 & $R_{o,2}$ & 30 & 0 & 4 & 0 & 25 & 1 \\
 & $R_{o,stress}$ & 9 & 9 & 0 & 0 & 0 & 0 \\
 & \emph{total} & \textbf{69} & \textbf{34} & \textbf{6} & \textbf{0} & \textbf{28} & \textbf{1} \\
\midrule
Spatial & $R_{s,1}$ & 30 & 24 & 0 & 0 & 4 & 2 \\
 & $R_{s,2}$ & 60 & 1 & 7 & 0 & 47 & 5 \\
 & $R_{s,3}$ & 30 & 30 & 0 & 0 & 0 & 0 \\
 & \emph{total} & \textbf{120} & \textbf{55} & \textbf{7} & \textbf{0} & \textbf{51} & \textbf{7} \\
\midrule
\multicolumn{2}{l}{\textbf{All}} & \textbf{381} & \textbf{200} & \textbf{59} & \textbf{5} & \textbf{101} & \textbf{16} \\
\bottomrule
\end{tabular}
\end{table}

Rule identifiers are suite-local ($R_{g,n}$ for Goal, $R_{10,n}$ for Long, $R_{o,n}$ for Object, $R_{s,n}$ for Spatial in \Cref{tab:app_perturbation_rules}), so the same number denotes different perturbations in different suites. Some generated cases were adjusted or dropped by hand to keep the scenes executable, so the rule IDs are coarse provenance groups, not exact templates. On Goal, an automatic nearest-neighbor check agrees with most human labels (\Cref{tab:goal_trajectory_nn}).

\begin{table}[htbp]
\centering
\scriptsize
\setlength{\tabcolsep}{5pt}
\caption{\textbf{Whole-trajectory nearest-neighbor check on Goal.} Goal rollouts labeled as retrieving the source or another task, with cached trajectories. A rollout is label-consistent when its nearest expert demonstration belongs to the task the human label names.}
\label{tab:goal_trajectory_nn}
\begin{tabular}{lcc}
\toprule
Trajectory metric & Usable rows & Label-consistent NN \\
\midrule
Object+EEF DTW & 99 & 87/99 (87.9\%) \\
EEF+gripper resampled distance & 99 & 83/99 (83.8\%) \\
\bottomrule
\end{tabular}
\end{table}

\subsection{Selected behavioral case studies}
\label{app:human_case_studies}

\paragraph{Whole-task switching.}
Under Goal $R_{g,4}$, which changes several task fields at once, 21 of 30 rollouts execute another familiar demonstrated task. Language blindness would predict that the policy mostly stays on the source trajectory. Instead, the modified instruction selects a different demonstrated-task basin. A changed destination on Spatial ($R_{s,3}$) keeps all 30 rollouts on the source placement trajectory.

\begin{figure*}[htbp]
    \centering
    \includegraphics[width=0.95\textwidth]{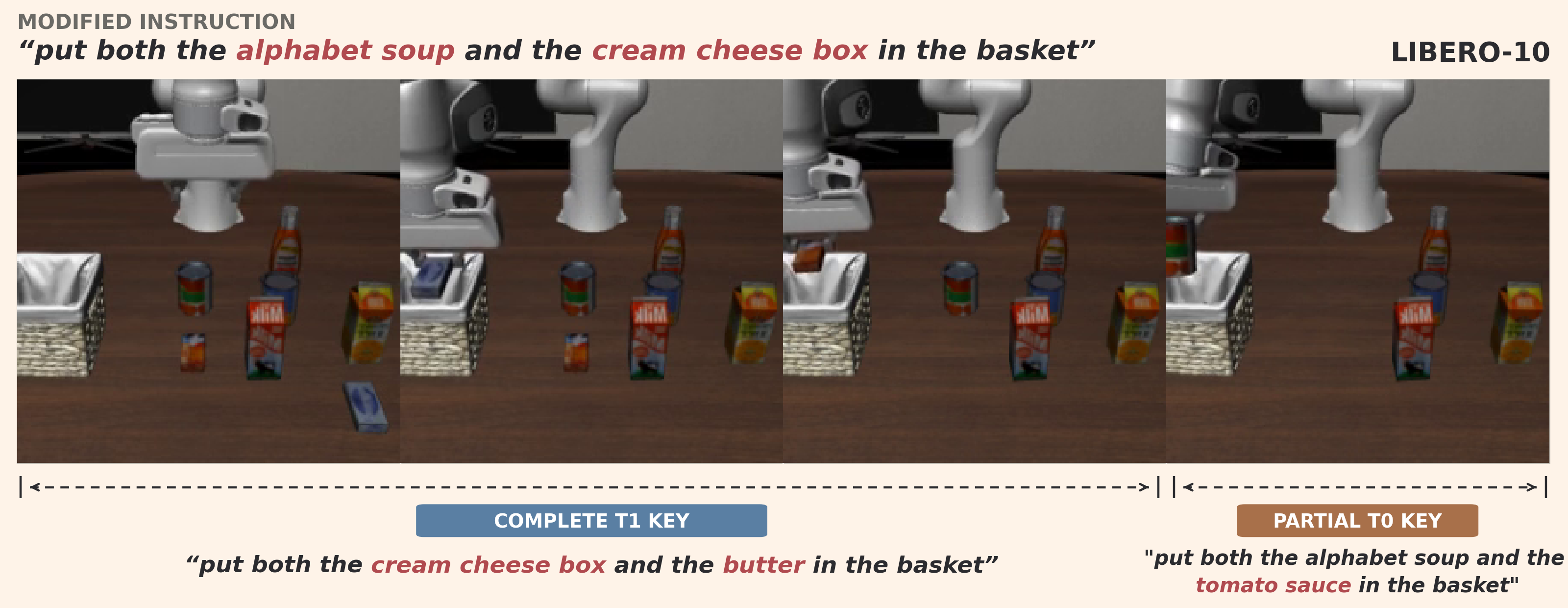}
    \vspace{-1.0em}
\caption{\textbf{A hybrid rollout combining familiar task segments.}
Under a modified two-object instruction, $\pi_{0.5}$ combines a complete segment associated with demonstrated task $T_1$ with a partial segment associated with task $T_0$.}\label{fig:libero10_hybrid_example}
\end{figure*}

\paragraph{Sub-trajectory retrieval.}
Hybrid rollouts can combine familiar segments from different demonstrated tasks. In \Cref{fig:libero10_hybrid_example}, $\pi_{0.5}$ follows a complete segment associated with task $T_1$ and then begins a segment associated with task $T_0$, instead of executing the requested two-object combination. This coherent sequence illustrates retrieval-like behavior at the level of sub-trajectories.

\paragraph{Object stress test with unchanged language.}
This probe separates choosing the object from tracking it (\Cref{tab:object_stress_test}). Under the unchanged instruction, we swap the instruction-named object with another object occupying the source-bound pickup slot and then shift the swapped object by 10~cm. All 9 rollouts pick the object in the source-bound slot rather than the named one and track it smoothly to its shifted position. The unchanged instruction isolates the scene intervention. Following the displacement also distinguishes this behavior from fixed-coordinate replay.

\begin{table}[t]
\centering
\scriptsize
\setlength{\tabcolsep}{4.0pt}
\caption{\textbf{Object stress test with unchanged language.}
The instruction is unchanged. The object occupying the source-bound slot is swapped and shifted by 10~cm.}
\label{tab:object_stress_test}
\resizebox{\ifdim\width>\linewidth\linewidth\else\width\fi}{!}{%
\begin{tabular}{lcl}
\toprule
Condition & $N$ & Observed behavior \\
\midrule
Same instruction, swapped object + 10 cm shift
& 9
& 9 / 9 follow the swapped-and-shifted source-bound object \\
\bottomrule
\end{tabular}%
}
\end{table}

\begin{table}[t]
\centering
\scriptsize
\setlength{\tabcolsep}{4pt}
\caption{\textbf{Perturbation rules for each LIBERO suite.}
Rule IDs match the row labels in \Cref{fig:app_all_rollouts}. Object rules leave the instruction unchanged and perturb only the scene.}
\label{tab:app_perturbation_rules}
\begin{tabular}{llp{0.58\linewidth}}
\toprule
Suite & Rule & Description \\
\midrule
Spatial & R1 & Replace the target object \\
                         & R2 & Alter the spatial description of the source location (object repositioned to match), with two variants \\
                         & R3 & Replace the placement destination \\
\midrule
Object  & R1     & Swap the target object's position with another in-scene object \\
                         & R2     & Slightly shift the target object's position \\
                         & stress & Swap and shift simultaneously \\
\midrule
Goal    & R1 & Change the action verb \\
                         & R2 & Replace the manipulated object \\
                         & R3 & Change the placement destination \\
                         & R4 & Combine two or more of the above modifications \\
                         & R5 & Truncate the instruction to its first sub-task \\
\midrule
Long      & R1 & Replace one of the two target objects \\
                         & R2 & Change the placement destination \\
                         & R3 & Replace the instruction entirely with one from a different task \\
\bottomrule
\end{tabular}
\end{table}

\begin{figure}[h]
    \centering
    \includegraphics[width=\linewidth]{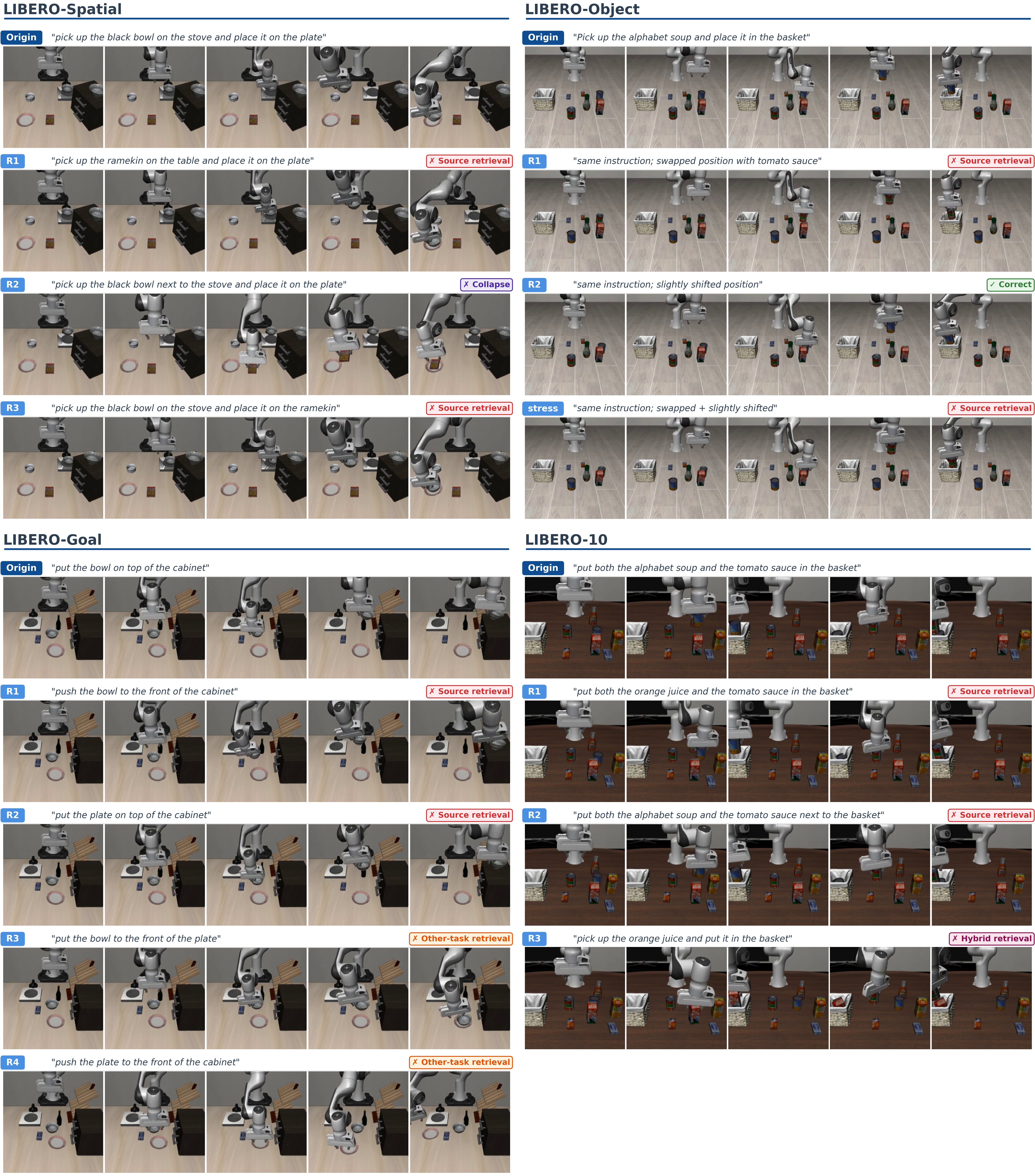}
\caption{\textbf{Rollout examples for all four LIBERO suites.}
Each panel shows the source behavior (Origin), followed by representative rollouts under the perturbation rules of \Cref{tab:app_perturbation_rules}, with outcome labels from \Cref{tab:human_label_taxonomy}.}    \label{fig:app_all_rollouts}
\end{figure}

\subsection{The same probes after ECT}
\label{app:probe_retrieval_after_ect}

To test whether ECT changes the diagnosed behavior itself, we reran the Goal and Spatial probes with the official rollouts' seeds and scenes on $\pi_{0.5}$ trained with ECT data + ECT loss under the Frozen-LM recipe (seed 42 in \Cref{tab:pi05_expertonly_seed_values}) and scored each rollout with two automatic measures (\Cref{tab:app_probe_retrieval_after_ect}). NN source uses the EEF+gripper resampled distance of \Cref{tab:goal_trajectory_nn}. On the official policy's Goal probes it marks 59 of 117 rollouts as source, where the human labels mark 52. Both policies complete the unperturbed source tasks at 95\% or more (\Cref{tab:libero_pro_diagnosis,tab:pi05_expertonly_seed_values}), so the reduced source-completion rate under perturbation is not accompanied by a general loss of source-task competence. The Goal set differs from \Cref{tab:app_human_rule_counts} in three rollouts.

\begin{table}[t]
\centering
\scriptsize
\setlength{\tabcolsep}{4pt}
\caption{\textbf{Binding probes before and after ECT.} Source completion measures how often the original task's predicate fires under a perturbed instruction. NN source measures how often the nearest expert demonstration belongs to the source task. Values are percentages of rollouts for one ECT seed.}
\label{tab:app_probe_retrieval_after_ect}
\begin{tabular}{llrrrrr}
\toprule
& & & \multicolumn{2}{c}{Source completion} & \multicolumn{2}{c}{NN source} \\
\cmidrule(lr){4-5}\cmidrule(lr){6-7}
Suite & Rule & $N$ & Official & ECT & Official & ECT \\
\midrule
Goal & $R_{g,1}$ & 33 & 84.8 & 66.7 & 90.9 & 78.8 \\
 & $R_{g,2}$ & 33 & 39.4 & 9.1 & 45.5 & 36.4 \\
 & $R_{g,3}$ & 21 & 38.1 & 19.0 & 38.1 & 19.0 \\
 & $R_{g,4}$ & 30 & 16.7 & 0.0 & 20.0 & 3.3 \\
 & \emph{total} & \textbf{117} & 46.2 & 24.8 & 50.4 & 36.8 \\
\midrule
Spatial & $R_{s,1}$ & 30 & 73.3 & 70.0 & 80.0 & 80.0 \\
 & $R_{s,2}$ & 30 & 70.0 & 90.0 & 73.3 & 93.3 \\
 & $R_{s,3}$ & 30 & 96.7 & 10.0 & 100.0 & 33.3 \\
 & \emph{total} & \textbf{90} & 80.0 & 56.7 & 84.4 & 68.9 \\
\bottomrule
\end{tabular}
\end{table}

On Goal, both measures fall on every rule, and under the combined perturbation $R_{g,4}$ the ECT policy never completes the source task. On Spatial, the change is concentrated where the instruction asks for a different action. Under a new destination ($R_{s,3}$), the official policy completes the original placement in 29 of 30 rollouts and the ECT policy in 3 of 30. Under a new object name ($R_{s,1}$), both policies still pick up the bowl. This pattern is consistent with the ECT data. Through Goal, the four-suite training set shows the same destination word at different places, but every Spatial instruction names the same black bowl, so nothing in the data asks the policy to read the object word. $R_{s,2}$ rewrites the spatial relation in the location clause, and the bowl on the plate can still satisfy the source predicate, so completion there does not separate following from retrieval.

Human labels on the full 381 probes agree (\Cref{tab:threeway_human}). The ECT policy carries out the changed task in 47.5\% of rollouts, against 26.5\% for the official $\pi_{0.5}$, with a gain on every suite. ECT therefore changes the behavior the probes measure, not only the counterfactual scores.

\begin{table}[t]
\centering
\scriptsize
\setlength{\tabcolsep}{3.5pt}
\caption{\textbf{Matched human labels for three policies on the 381 probes.} Follow denotes correct or near-correct execution of the changed task. Retr.\ denotes source-task, other-task, or hybrid retrieval. Coll.\ denotes behavior without a coherent trajectory. Values are percentages of rollouts. ECT is the Frozen-LM $\pi_{0.5}$ model trained with ECT data + ECT loss (seed 42).}
\label{tab:threeway_human}
\begin{tabular}{@{}lr ccc ccc ccc@{}}
\toprule
& & \multicolumn{3}{c}{Official $\pi_{0.5}$} & \multicolumn{3}{c}{Official GR00T-N1.7} & \multicolumn{3}{c}{$\pi_{0.5}$ + ECT} \\
\cmidrule(lr){3-5}\cmidrule(lr){6-8}\cmidrule(l){9-11}
Suite & $N$ & Follow & Retr. & Coll. & Follow & Retr. & Coll. & Follow & Retr. & Coll. \\
\midrule
Spatial & 120 & 42.5 & 51.7 & 5.8 & 33.3 & 54.2 & 12.5 & \textbf{64.2} & 31.7 & 4.2 \\
Object & 69 & 40.6 & 58.0 & 1.4 & 15.9 & 84.1 & 0.0 & \textbf{75.4} & 24.6 & 0.0 \\
Goal & 120 & 12.5 & 82.5 & 5.0 & 0.8 & 71.7 & 27.5 & \textbf{26.7} & 60.8 & 12.5 \\
Long & 72 & 9.7 & 87.5 & 2.8 & 0.0 & 97.2 & 2.8 & \textbf{27.8} & 72.2 & 0.0 \\
\midrule
All & 381 & 26.5 & 69.3 & 4.2 & 13.6 & 73.2 & 13.1 & \textbf{47.5} & 47.2 & 5.2 \\
\bottomrule
\end{tabular}
\end{table}
\section{Mechanistic Diagnostic Details}
\label{app:mechanistic_details}

This appendix details the prefix-KV retrieval, patching, and persistent intervention of \Cref{sec:mechanistic_evidence}.

\subsection{Prefix-KV retrieval protocol and robustness}
\label{app:kv_analysis}
\label{app:kv_robustness}

\begin{figure}[t]
    \centering
    \includegraphics[width=0.8\linewidth]{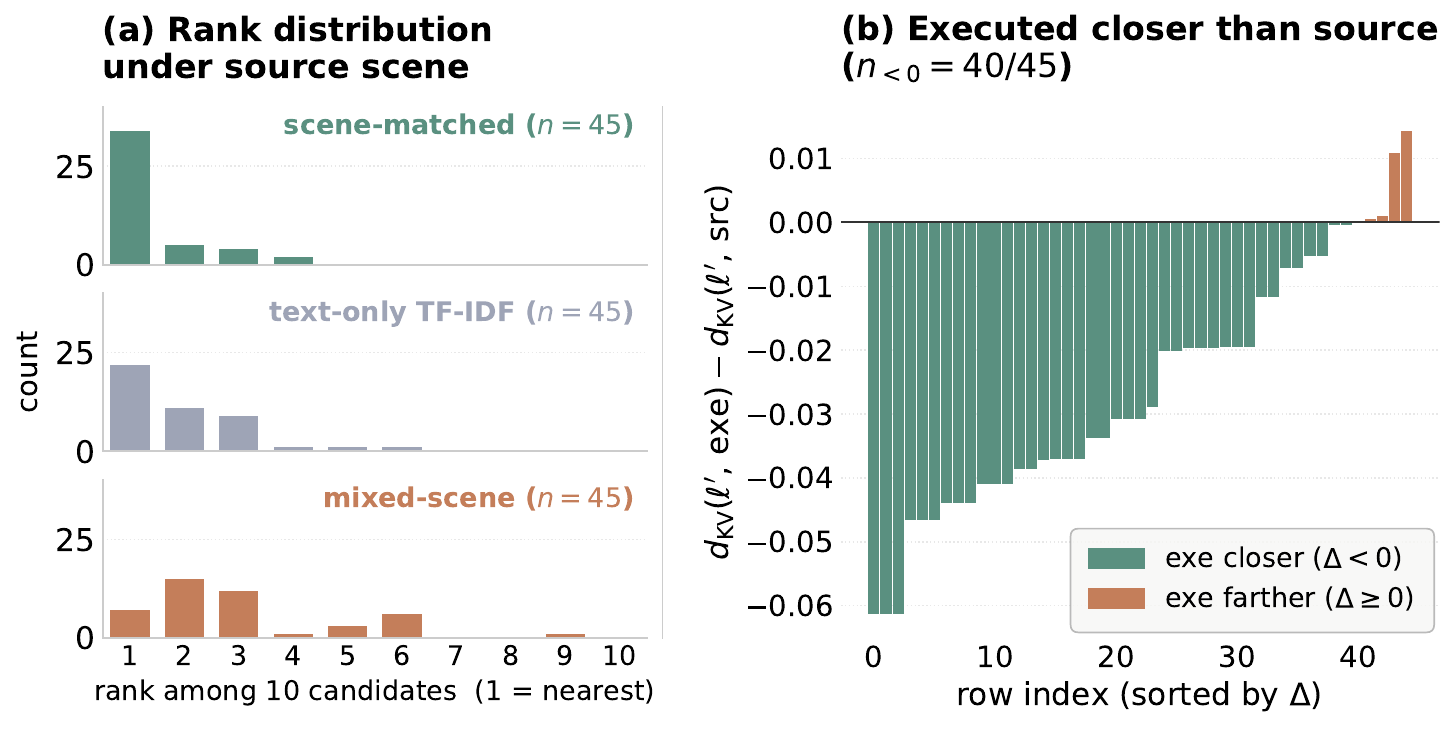}
    \caption{\textbf{Executed tasks are recoverable from scene-matched prefix-KV states.} \textbf{(a)} Rank of the executed task among the ten candidates, for scene-matched prefix-KV states, a text-only TF-IDF baseline over the instructions, and a mixed-scene prefix-KV control. \textbf{(b)} The perturbed input is closer to the executed task than to the source task in scene-matched prefix-KV space.}
    \label{fig:kv_executed_task}
\end{figure}

\begin{figure}[htbp]
    \centering
    \includegraphics[width=\linewidth]{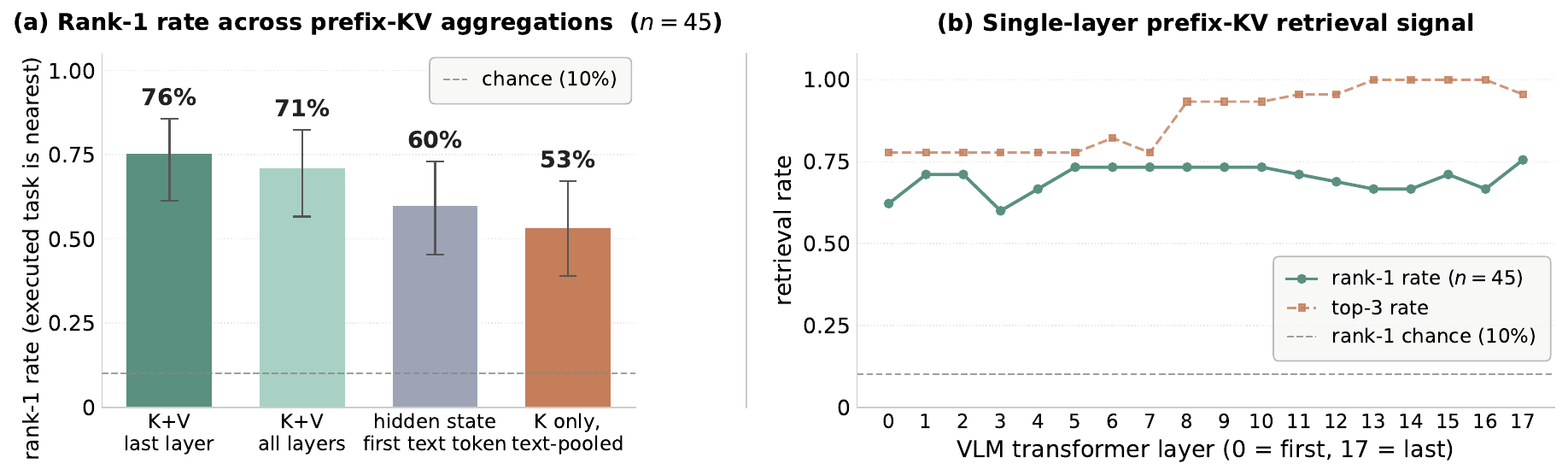}
    \caption{\textbf{Prefix-KV retrieval is robust to representation aggregation.} Same 45 rollouts as \Cref{fig:kv_executed_task}. \textbf{(a)} Rank-1 recovery under alternative representations. Error bars are 95\% Wilson intervals, and the dashed line marks chance. \textbf{(b)} Single-layer prefix K+V retrieval by VLM layer.}
    \label{fig:app_kv_robustness}
\end{figure}

For each input we extract the key and value tensors that the VLM expert produces at the visual-language prefix positions. This cache is not the action expert's own state, which consists of separate suffix-token states, but the prefix that its tokens attend to during action generation. We call it the \emph{prefix-KV representation}. With $K^{(L)}(s,\ell)$ and $V^{(L)}(s,\ell)$ the final-layer keys and values at prefix positions for scene $s$ and instruction $\ell$, the main-text representation is
\begin{equation}
\phi_{\mathrm{KV}}(s,\ell)
=
\frac{
\operatorname{vec}\!\left[K^{(L)}(s,\ell), V^{(L)}(s,\ell)\right]
}{
\left\|
\operatorname{vec}\!\left[K^{(L)}(s,\ell), V^{(L)}(s,\ell)\right]
\right\|_2
},
\label{eq:app_kv_representation}
\end{equation}
where $\operatorname{vec}[\cdot]$ flattens the final-layer prefix K+V tensors, with cosine distance
\begin{equation}
d_{\mathrm{KV}}(x,y)
=
1-\phi_{\mathrm{KV}}(x)^\top \phi_{\mathrm{KV}}(y).
\label{eq:app_kv_distance}
\end{equation}
We use the final layer by default. \Cref{fig:app_kv_robustness} reports other layers and aggregations.

The queries are a fixed set of 45 Goal rollouts labeled \texttt{same\_to\_other} when this analysis was run, each executing another familiar demonstrated task. For each query we keep the source scene fixed and compare the perturbed input with the ten candidate task instructions under that scene,
\begin{equation}
q=(s_{\mathrm{src}},\ell_{\mathrm{pert}}),
\qquad
x_c=(s_{\mathrm{src}},\ell_c),
\qquad
c\in\{1,\ldots,10\},
\label{eq:app_kv_scene_matched_inputs}
\end{equation}
and rank the executed task by
\begin{equation}
\operatorname{rank}_{\mathrm{exe}}(q)
=
1+
\sum_{c\neq c_{\mathrm{exe}}}
\mathbf{1}
\!\left[
d_{\mathrm{KV}}(q,x_c)
<
d_{\mathrm{KV}}(q,x_{c_{\mathrm{exe}}})
\right].
\label{eq:app_kv_rank}
\end{equation}
The executed task is the nearest candidate in 34 of 45 rollouts, against 7 of 45 in a mixed-scene control, and is usually closer than the source task (\Cref{fig:kv_executed_task}). The signal does not depend on how the representation is aggregated (\Cref{fig:app_kv_robustness}). All-layer K+V, last-layer hidden states, and K-only text pooling all preserve it, and each layer reaches at least 60\% rank-1 recovery with a high top-3 rate.

\subsection{Prefix-KV patching intervention}
\label{app:kv_patching}

Retrieval is correlational. To test whether the prefix affects action decoding, we replace the prefix K/V cache of each of the 45 perturbed inputs with the cache of a candidate task under the same source scene and decode the action chunk with the action expert unchanged. Let $A_{\mathrm{nat}}$ and $A_{\mathrm{patch}}$ denote the natural and patched decodes, and $A_T$ the expert chunk for task $T$. The improvement
\begin{equation}
\Delta_{\mathrm{L2}}(T)
=
\left\|A_{\mathrm{nat}}-A_T\right\|_2
-
\left\|A_{\mathrm{patch}}-A_T\right\|_2
\label{eq:kv_patch_l2_improvement}
\end{equation}
is positive when patching moves the decode closer to the target expert. We report it for the first step and the full chunk (\Cref{tab:app_kv_patching}).

\begin{table}[htbp]
\centering
\scriptsize
\setlength{\tabcolsep}{4pt}
\caption{\textbf{Prefix-KV patching intervention.}
Improvement $\Delta_{\mathrm{L2}}$ (\Cref{eq:kv_patch_l2_improvement}) with bootstrap intervals over the 45 rollouts. Positive values mean the patched decode is closer to the target expert than the natural perturbed decode. Pos.\ frac.\ is the fraction of rollouts with positive improvement.}
\label{tab:app_kv_patching}
\begin{tabular}{llrrrr}
\toprule
Patch & Target & First L2 impr. & Pos. frac. & Chunk L2 impr. & Pos. frac. \\
\midrule
Source final & Source & $0.0024$ $[0.0011,0.0036]$ & 0.69 & $0.0182$ $[0.0107,0.0262]$ & 0.87 \\
Executed final & Executed & $0.0009$ $[-0.0004,0.0022]$ & 0.62 & $0.0021$ $[0.0006,0.0035]$ & 0.62 \\
Source all & Source & $-0.0208$ $[-0.0418,-0.0012]$ & 0.42 & $0.5467$ $[0.3953,0.6949]$ & 0.84 \\
Executed all & Executed & $-0.0094$ $[-0.0237,0.0045]$ & 0.42 & $0.1048$ $[0.0356,0.1786]$ & 0.62 \\
\bottomrule
\end{tabular}
\end{table}

Restoring the source-task prefix tests whether the decode can be pulled back from the executed task toward the source basin, and the chunk-level distance to the source expert improves in 87\% of rollouts, with an interval excluding zero. Executed-task patching has a weaker effect because the natural decode is already close to the executed basin. All-layer patching changes the chunk more but worsens the first-step distance, showing that a larger intervention need not improve both measures.

\paragraph{Target specificity.}
For each rollout we patch each of the ten candidate prefixes, all extracted under the source-scene observation, and compare the patch-induced action displacement with the patched candidate's canonical LIBERO expert chunk, matched by episode index. This measures steering toward a candidate task's expert trajectory. Own-target alignment exceeds off-target alignment under all three measures (\Cref{tab:kv_patch_directional}). Source-prefix patches also project further toward the source expert than the mean-other and farthest-other prefixes (paired differences of 0.006 and 0.010, bootstrap intervals excluding zero), showing that different valid prefixes do not have interchangeable effects. The persistent intervention below tests whether repeated replacement redirects the rollout.

\begin{table}[htbp]
\centering
\scriptsize
\setlength{\tabcolsep}{5pt}
\caption{\textbf{Target specificity of prefix-KV patching.} Own target measures alignment of the patch-induced displacement with the patched candidate's expert trajectory. Off-target averages alignment with the other candidates. Specificity is their difference. Brackets report rollout-bootstrap intervals.}
\label{tab:kv_patch_directional}
\begin{tabular}{lccc}
\toprule
Analysis & Own target & Off-target & Specificity \\
\midrule
Cosine alignment & $0.443$ $[0.400,0.487]$ & $0.182$ $[0.148,0.217]$ & $0.262$ $[0.234,0.288]$ \\
Normalized projection & $0.0136$ $[0.0102,0.0174]$ & $0.0064$ $[0.0044,0.0085]$ & $0.0073$ $[0.0054,0.0092]$ \\
Raw projection & $0.0215$ $[0.0160,0.0275]$ & $0.0089$ $[0.0061,0.0120]$ & $0.0125$ $[0.0096,0.0158]$ \\
\bottomrule
\end{tabular}
\end{table}

\paragraph{Persistent (rollout-level) protocol.}
The persistent intervention of \Cref{sec:kv_retrieval} replaces the prefix-KV cache at every policy call rather than at one. Source scene, source instruction, and policy are fixed. The injected cache is computed once from the target task's initial observation and instruction. Controls keep the source cache. The 50 pairs, enumerated before any rollout, are all ordered (source, target) pairs of the Object suite whose target object is present in the source scene, with four rollouts per condition. For the official GR00T-N1.7 Object checkpoint, which has no prefix-KV cache, we replace the backbone feature sequence that its action head cross-attends to (a function of pixels and instruction only) under the same protocol.

Both architectures show task-directed redirection (\Cref{tab:app_persistent_intervention}). Injection moves the arm nearer the target task's object in over half of the rollouts and in no control, with a larger shift than the matched control in every pair. Outcomes are often all or nothing within a pair. No injected rollout completes the target task. The replaced representation is fixed at the target task's initial observation rather than updated from the live scene. The pairs that resist redirection are largely shared, with 14 of the 19 for GR00T-N1.7 also resisting it for $\pi_{0.5}$.

\begin{table}[htbp]
\centering
\scriptsize
\setlength{\tabcolsep}{4pt}
\caption{\textbf{Persistent intervention on Object, full pair grid.} Four rollouts per condition cover all 50 valid ordered (source, target) pairs. Closer means the end effector is nearer the target than the source object. Larger shift means greater movement toward the target under injection than control. Under injection, source-task completion is 0/200 for both models.}
\label{tab:app_persistent_intervention}
\resizebox{\ifdim\width>\linewidth\linewidth\else\width\fi}{!}{%
\begin{tabular}{lrrrrrrr}
\toprule
Model & Injected closer & Control closer & Pairs, larger shift & Pairs 4/4 & Pairs 0/4 & Control completes source & Injected completes target \\
\midrule
$\pi_{0.5}$ (prefix-KV) & 109/200 & 0/200 & 50/50 & 25 & 21 & 200/200 & 0/200 \\
GR00T-N1.7 (backbone features) & 121/200 & 0/200 & 50/50 & 28 & 19 & 200/200 & 0/200 \\
\bottomrule
\end{tabular}%
}
\end{table}

\paragraph{Real-robot offline decomposition.}
For three UR5e models (a Standard model trained on a quarter of the demonstrations, ECT data, and ECT data + ECT loss), we patch offline on recorded frames with images and proprioception held fixed, swapping either only the instruction or only the vision inside the cache. In all three models, swapping the instruction retargets the predicted chunk in about three quarters of 48 pairs, against about one fifth for swapping the vision. Instruction-dependent retargeting therefore persists after ECT. The measured outcome is instruction-dependent retargeting in the recorded frames.

\section{ECT Data Construction}
\label{app:ect_details}

This appendix details the ECT data construction (\Cref{sec:ect_method}) and compares the transformed layouts with LIBERO-PRO Swap evaluation layouts.

\subsection{Transformation operators and action validity}
\label{app:ect_transform_details}

\begin{figure}[htbp]
      \centering
      \includegraphics[width=\linewidth]{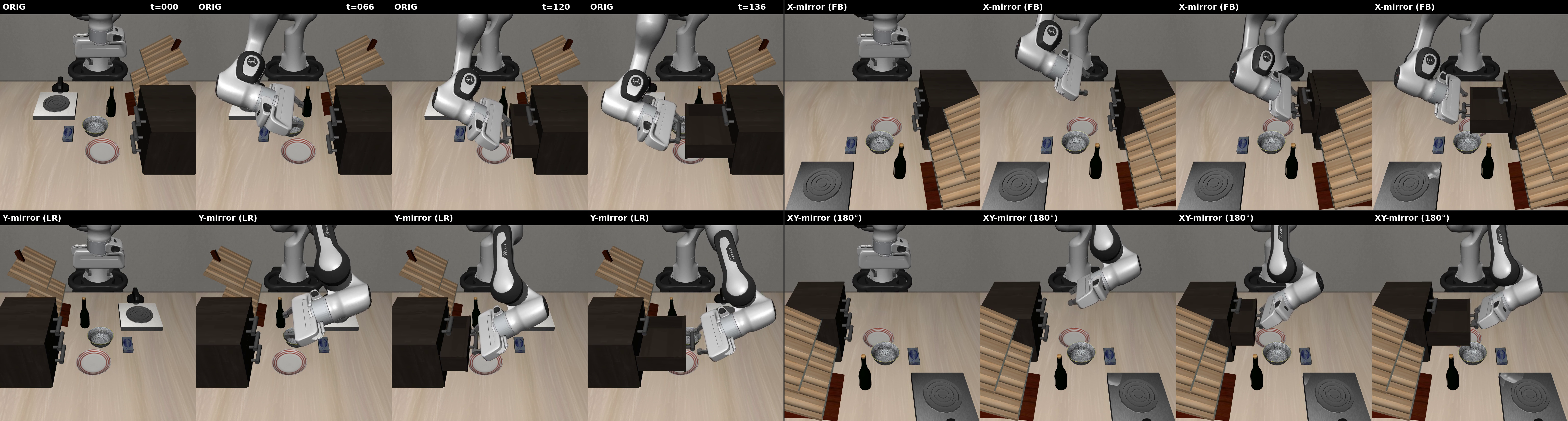}
      \caption{\textbf{ECT scene and trajectory transformations.}
      The top row shows the original demonstration and x-axis mirror (front-back).
      The bottom row shows the y-axis mirror (left-right) and xy-axis mirror ($180^\circ$ rotation).
      Each panel shows four timesteps. The robot is unchanged.}
      \label{fig:ect_transforms}
\end{figure}

For constructed pairs, ECT uses curated scene changes rather than discovering symmetries. The alternatives are intended to expose scene-dependent action choices under a fixed instruction. We write $(\tilde{s},\ell,\tilde{a})=(s',\ell,a')$ for the constructed branch of the five-tuple in \Cref{eq:ect_tuple}. For each original demonstration $(s,\ell,a)$, the builder constructs
\begin{equation}
(\tilde{s},\ell,\tilde{a})
=
\mathcal{T}_{\mathrm{ECT}}(s,\ell,a),
\qquad
\tilde{s}=M_s(s),
\quad
\tilde{a}=\mathrm{Replay}_{\tilde{s}}\!\left(M_a(a)\right),
\end{equation}
with $\mathrm{Replay}_{\tilde{s}}$ as in \Cref{eq:ect_two_branch}. \emph{Mirror} transforms reflect the scene about a workspace axis (\Cref{fig:ect_transforms}). \emph{Shift} transforms translate selected scene elements and are used where a pure mirror would create visually ambiguous layouts or unstable transformed scenes. \Cref{tab:ect_transform_provenance} lists the family used per suite.

\paragraph{Label provenance.}
The builder transforms the end-effector waypoint path and the rotation command analytically and keeps the gripper command. It then replays the demonstration in the transformed scene with a tracking controller and records the \emph{achieved} end-effector displacement as the translational action, while the original branch stores the commanded action. Because the robot base is never mirrored, the arm reaches mirrored waypoints through a different configuration. On the y-mirror pairs of Spatial, the rotation and gripper channels equal the reflected command in every frame, while the translation differs from it by a median of about $0.1$ on the $[-1,1]$ action scale. ``Action-valid'' thus means that $\tilde a$ is a kinematically valid expert action for $\tilde s$, verified by task success, not a symmetry image of $a$, and ECT is not an equivariance constraint on actions.

\begin{figure}[t]
    \centering
    \includegraphics[width=0.85\linewidth]{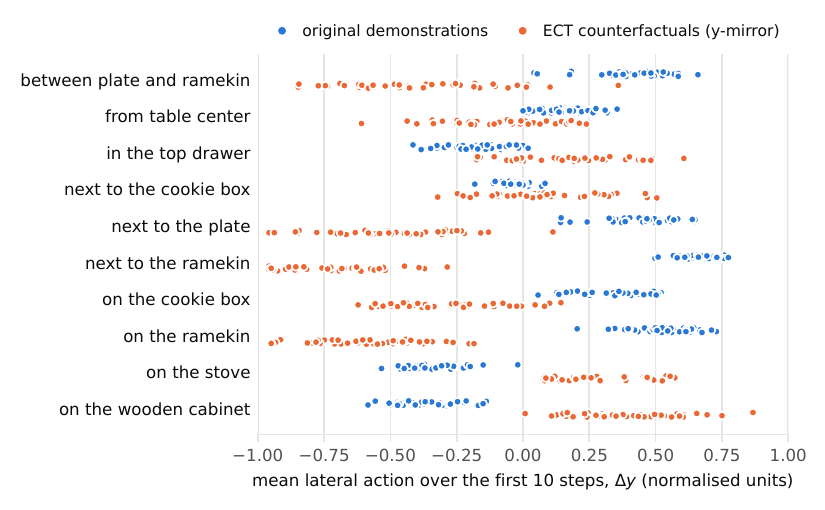}
    \caption{\textbf{What ECT does to conditional action support (Spatial).} Each dot is a demonstration's mean lateral action over its first ten steps, for original data (blue) or y-mirror counterfactuals (orange).}
    \label{fig:conditional_support}
\end{figure}

In the original Spatial data each instruction's action support lies on one side, as the bowl is always reached from one direction (\Cref{fig:conditional_support}). The y-mirror counterfactuals put the same instruction on the other side, so a fixed instruction-to-family assignment no longer matches the demonstrated scene-dependent choices (\Cref{sec:conditional_support}). The replayed translation widens their spread.

\paragraph{Design principles.}
The construction makes the instruction insufficient to select the appropriate branch while retaining observations that distinguish it.
(i)~\emph{Same instruction, different required action.} Changing the relevant scene geometry gives a different action under the same instruction. The purpose is scene-dependent action choice, not arbitrary variability among equally valid trajectories (\Cref{fig:conditional_support}).
(ii)~\emph{Executable labels.} In the synthetic branch, transformed paths are replayed and filtered by task success. The rotation, translation, and gripper-label conventions are described above.
(iii)~\emph{Visual separability.} The scene must reveal which action is appropriate. The Object construction in \Cref{fig:ect_obj_shift_comp} gives the counterexample. Front-back mirroring keeps dominant landmarks in similar image regions while changing the required action. Left-right mirroring or a post-mirror shift moves those landmarks and restores scene-level distinguishability.

\begin{table}[t]
\centering
\scriptsize
\setlength{\tabcolsep}{5pt}
\caption{\textbf{Object-suite construction diagnostics.} Success (\%) for per-suite $\pi_{0.5}$ LoRA diagnostics, separate from the main Frozen-LM comparison. X-mirror and all-mirror variants lose ID and Object competence, while Y-mirror and shifted variants preserve them.}
\label{tab:object_transform_diagnostics}
\begin{tabular}{lrrrr}
\toprule
Construction & ID & Sem. & Obj. & Swap \\
\midrule
No ECT baseline & 94.2 & 94.8 & 84.4 & 0.0 \\
X-mirror, paired & 23.8 & 22.6 & 25.4 & 3.0 \\
Y-mirror, paired & 93.2 & 96.4 & 85.0 & 1.0 \\
All mirrors, paired & 29.0 & 21.8 & 20.2 & 12.0 \\
X-mirror + X-shift, paired & 97.4 & 99.2 & 95.8 & 1.6 \\
XY-mirror + XY-shift, paired & 95.6 & 98.6 & 91.4 & 7.6 \\
\bottomrule
\end{tabular}
\end{table}

X-mirror and all-mirror banks strongly reduce nominal competence on Object, whereas Y-mirror and the two shifted constructions preserve ID, Semantic, and Object success. These diagnostics preceded the four-suite training runs. The Object ECT data combine exactly these three constructions and exclude the all-mirror bank, although it has the highest Swap in the table, so the choice follows nominal competence rather than Swap. Swap stays low in these per-suite runs, so separability makes the counterfactual supervision learnable but does not by itself raise Object Swap. The four-suite results are in \Cref{tab:counterfactual_main,tab:app_pi05_unified_ect}.

\begin{table}[htbp]
\centering
\scriptsize
\setlength{\tabcolsep}{3pt}
\caption{\textbf{ECT transform provenance.}
Transform choices are fixed before training by the action-validity and visual-separability criteria, not selected per evaluation rollout.}
\label{tab:ect_transform_provenance}
\resizebox{\ifdim\width>\linewidth\linewidth\else\width\fi}{!}{%
\begin{tabular}{llll}
\toprule
Suite & Transform family & Action transform & Role in ECT \\
\midrule
Spatial & mirror & reflected, then replayed & changes placement direction under fixed instruction \\
Object & Y-mirror / mirror+shift & reflected (x and xy also shifted), then replayed & preserves separability while changing source-object branch \\
Goal & mirror & reflected, then replayed & changes goal-relative spatial branch \\
Long & shift-augmented mix & shifted, then replayed & preserves long-horizon executability \\
\bottomrule
\end{tabular}%
}
\end{table}

\paragraph{Contrast with prior equivariance-based augmentation.}
RoCoDA \citep{ameperosa2025rocoda} combines causal resampling with $SE(3)$ transformations and visual augmentation, evaluating ACT in a single-task setting. It saves successful transformed sub-trajectories. ECT shares the use of action-valid transformed demonstrations, but targets instruction-action binding in VLA fine-tuning. Its paired component preserves the correspondence between scene-dependent alternatives under the same instruction. \Cref{tab:ect_decomposition} compares this procedure with independent sampling of the same counterfactual pool. Natural CALVIN pairs and physically collected UR5e alternatives apply it without the LIBERO construction.

\paragraph{Use in training and scope.}
ECT data enter the same imitation-learning pipeline as the original demonstrations. The standard loss samples them independently of their originals. The ECT loss places each original and its counterpart, which share the instruction but not the action, in the same update (\Cref{eq:ect_pair_loss}). In our $\pi_{0.5}$ implementation the two halves of a pair also share the flow-matching noise sample and flow time, while image augmentations are drawn independently for each half. Same-instruction spatial counterfactuals directly target Swap, which requires selecting the correct spatial branch under a fixed task identity. Task perturbations instead change the referred object, goal, relation, or action. ECT does not construct such counterfactuals, so its Task gains (\Cref{tab:counterfactual_main}) provide complementary evidence of transfer beyond the scene changes directly constructed during training.

\subsection{ECT counterfactuals are built independently of LIBERO-PRO evaluation}
\label{app:ect_no_leakage}

ECT counterfactuals transform the scene and action of LIBERO training demonstrations $(s,\ell,a)$ together under a fixed instruction. LIBERO-PRO perturbations are generated independently at evaluation time by rewriting BDDL fields. Semantic paraphrases the instruction, object perturbation changes object attributes under a fixed instruction, Swap permutes object or destination positions in the initial state, and Task rewrites the instruction and goal block. The two constructions operate on different inputs, training trajectories versus evaluation BDDL specifications.

\paragraph{Layout distance.}
We compare Swap test states with original demonstration layouts and with the successful transformed layouts added by ECT. The transformed set excludes originals, although ECT models train on both. For each state, we take the nearest layout of the same task using the mean planar distance over its objects and fixtures. The ID reference scale is the 90th percentile of ID-to-original distances. No Swap state falls within this scale of the transformed set, whose smallest distance is 6.4~cm (\Cref{tab:swap_coverage}). Its median distance exceeds that of the originals in every suite. For the exchanged objects alone, one Spatial task has local overlap within the corresponding ID scale. These distances describe whole layouts rather than the novelty of each object's position.

\begin{table}[htbp]
\centering
\scriptsize
\setlength{\tabcolsep}{5pt}
\caption{\textbf{Layout distances from Swap states to original and transformed demonstrations.} Mean planar object-and-fixture distance (cm) to the nearest layout of the same task over 500 Swap states per suite. The transformed set contains successful counterfactual layouts only. ID scale is the 90th percentile of ID-to-original distances.}
\label{tab:swap_coverage}
\begin{tabular}{lccccc}
\toprule
 & & \multicolumn{2}{c}{Median distance to} & Minimum distance & Within ID scale \\
\cmidrule(lr){3-4}
Suite & ID scale & original & transformed & to transformed & of transformed \\
\midrule
Spatial & 1.0 & 6.6 & 28.7 & 13.3 & 0\% \\
Object & 0.3 & 6.4 & 28.5 & 25.2 & 0\% \\
Goal & 0.9 & 11.3 & 24.9 & 19.6 & 0\% \\
Long & 1.7 & 13.9 & 16.2 & 6.4 & 0\% \\
\bottomrule
\end{tabular}
\end{table}

The training-only construction inputs and the layout comparison address separation from the evaluation states. Broader distributional coverage can contribute to generalization without reproducing a complete test layout. Task evaluation also changes the instructions and goals held fixed by ECT.

\section{ECT Results on LIBERO}
\label{app:ect_libero_results}

This appendix reports the complete LIBERO-PRO results of the ECT models, the decomposition across seeds and suites, the data-scale sweep, and ECT under RL post-training.

\subsection{\texorpdfstring{$\pi_{0.5}$}{pi0.5} ECT results}
\label{app:pi05_ect_results}
\label{app:recipes}

\Cref{tab:app_pi05_unified_ect} reports all four-suite $\pi_{0.5}$ models. Its Frozen-LM rows are the three models of \Cref{tab:ect_decomposition}. Both recipes (\Cref{tab:recipes}) train the SigLIP vision encoder (ViT, 414.8M parameters) and the 300M action expert (AE, 427.9M measured) and differ in whether the PaliGemma language model (LM, 2{,}508.5M) is trained. Models compared within a recipe share one trainable set. ECT uses mirror-based constructions, with x-shift and xy-shift variants on Object and selected Long tasks to address ambiguous layouts. The suite-specific families are listed in \Cref{tab:ect_transform_provenance}. All cells use the evaluation protocol of \Cref{app:libero_pro_patch}.

\begin{table}[htbp]
\centering
\scriptsize
\setlength{\tabcolsep}{4pt}
\caption{\textbf{Training recipes by trainable parameter group ($\pi_{0.5}$, measured).} Millions of parameters that receive gradients under each recipe's trainable filter. Both recipes train the heads (2.2M).}
\label{tab:recipes}
\begin{tabular}{@{}llcccr@{}}
\toprule
Recipe & Configuration & LM & ViT & AE & Trainable \\
\midrule
Frozen LM & \texttt{\_expert\_only} & frozen  & trained & trained & 844.9M \\
Full FT   & \texttt{pi05\_libero} (official), \texttt{\_full\_sft} (ECT) & trained & trained & trained & 3{,}353.4M \\
\bottomrule
\end{tabular}
\end{table}

\begin{table}[t]
\centering
\scriptsize
\setlength{\tabcolsep}{3.2pt}
\caption{\textbf{Five-cell results of the four-suite $\pi_{0.5}$ models.} $N=500$ per cell and seed. Recipes follow \Cref{tab:recipes}. ECT rows shaded. $^\dagger$Mean over three seeds (0, 1, 42), with standard deviations of ECT data + ECT loss in \Cref{tab:pi05_expertonly_seed_values}.}
\label{tab:app_pi05_unified_ect}
\begin{tabular}{lllrrrrr}
\toprule
Suite & Model & Recipe & ID & Sem. & Obj. & Swap & Task \\
\midrule
Spatial & Official $\pi_{0.5}$ & Full FT & 98.4 & 98.0 & 98.0 & 46.6 & 53.0 \\
\rowcolor{teal!7}
 & ECT data + ECT loss & Full FT & 95.8 & 94.0 & 95.6 & 72.8 & 53.2 \\
 & Standard$^\dagger$ & Frozen LM & 95.9 & 94.5 & 95.4 & 52.7 & 22.3 \\
\rowcolor{teal!7}
 & ECT data$^\dagger$ & Frozen LM & 96.5 & 93.0 & 95.2 & 72.1 & 18.5 \\
\rowcolor{teal!7}
 & ECT data + ECT loss$^\dagger$ & Frozen LM & 98.4 & 96.1 & 97.1 & 74.4 & 22.1 \\
\midrule
Object & Official $\pi_{0.5}$ & Full FT & 98.6 & 99.0 & 94.4 & 18.2 & 11.0 \\
\rowcolor{teal!7}
 & ECT data + ECT loss & Full FT & 98.8 & 98.8 & 93.0 & 40.8 & 28.4 \\
 & Standard$^\dagger$ & Frozen LM & 97.1 & 98.0 & 83.5 & 38.3 & 23.0 \\
\rowcolor{teal!7}
 & ECT data$^\dagger$ & Frozen LM & 98.5 & 98.9 & 81.9 & 71.1 & 37.3 \\
\rowcolor{teal!7}
 & ECT data + ECT loss$^\dagger$ & Frozen LM & 99.1 & 99.2 & 78.5 & 70.8 & 31.3 \\
\midrule
Goal & Official $\pi_{0.5}$ & Full FT & 96.4 & 94.6 & 88.8 & 34.2 & 21.2 \\
\rowcolor{teal!7}
 & ECT data + ECT loss & Full FT & 94.6 & 94.2 & 87.2 & 35.8 & 26.6 \\
 & Standard$^\dagger$ & Frozen LM & 95.2 & 92.9 & 85.9 & 41.3 & 20.6 \\
\rowcolor{teal!7}
 & ECT data$^\dagger$ & Frozen LM & 93.8 & 90.0 & 82.8 & 51.8 & 40.7 \\
\rowcolor{teal!7}
 & ECT data + ECT loss$^\dagger$ & Frozen LM & 95.6 & 93.3 & 86.7 & 55.5 & 35.7 \\
\midrule
Long & Official $\pi_{0.5}$ & Full FT & 91.8 & 90.8 & 66.8 & 9.4 & 17.0 \\
\rowcolor{teal!7}
 & ECT data + ECT loss & Full FT & 92.0 & 92.6 & 62.4 & 26.6 & 26.6 \\
 & Standard$^\dagger$ & Frozen LM & 93.7 & 91.3 & 60.3 & 12.7 & 19.5 \\
\rowcolor{teal!7}
 & ECT data$^\dagger$ & Frozen LM & 92.9 & 93.0 & 56.7 & 29.5 & 22.3 \\
\rowcolor{teal!7}
 & ECT data + ECT loss$^\dagger$ & Frozen LM & 92.8 & 92.9 & 59.1 & 34.7 & 22.5 \\
\bottomrule
\end{tabular}
\end{table}

Under the Frozen-LM recipe, ECT data + ECT loss gives the highest Swap on three suites and the ECT data alone on Object, with ID and Semantic success close to their references. On Spatial, Task success is lower under the Frozen-LM recipe than under full fine-tuning for Standard and ECT alike, a difference associated with the fine-tuning recipe in these comparisons, rather than a unique effect of ECT. On the other three suites, Frozen-LM ECT exceeds the official $\pi_{0.5}$ on Task.

\subsection{GR00T-N1.7 ECT results}
\label{app:groot_ect_results}

We evaluate GR00T-N1.7 under per-suite and four-suite training. All GR00T-N1.7 ECT models train on the ECT data with the standard loss. The per-suite models use each suite's ECT data (x-shift on Object, shift-augmented on Long) and train for $30$K steps at batch size $32$. The four-suite model trains on the ECT data of all four suites for $20$K steps at a global batch size of $640$, matching the official GR00T-N1.7 LIBERO recipe ($20$K steps, batch $640$). The vision encoder and language model are frozen, while the projector and action head are trained.

\begin{table}[t]
\centering
\scriptsize
\setlength{\tabcolsep}{8pt}
\renewcommand{\arraystretch}{1.16}
\caption{\textbf{GR00T-N1.7 ECT full results.} Success (\%) over 500 trials per cell. All ECT models use the ECT data with the standard loss. ECT rows are shaded.}
\label{tab:app_groot_ect_full}
\begin{tabular}{lllrrrrr}
\toprule
Suite & Model & Training scope & ID & Sem. & Obj. & Swap & Task \\
\midrule
Spatial & Official & Per suite & 93.6 & 88.8 & 93.0 & 1.2 & 51.4 \\
\rowcolor{teal!7}  & ECT data & Per suite & 87.0 & 91.4 & 83.2 & 17.8 & 53.0 \\
\rowcolor{teal!7}  & ECT data & Four suites & 90.8 & 88.0 & 89.8 & 38.2 & 64.0 \\
\midrule
Object & Official & Per suite & 95.6 & 97.6 & 86.0 & 0.0 & 9.0 \\
\rowcolor{teal!7}  & ECT data & Per suite & 97.6 & 98.2 & 88.6 & 0.0 & 10.0 \\
\rowcolor{teal!7}  & ECT data & Four suites & 97.2 & 97.2 & 79.8 & 7.0 & 10.2 \\
\midrule
Goal & Official & Per suite & 94.2 & 93.8 & 76.0 & 2.2 & 10.0 \\
\rowcolor{teal!7}  & ECT data & Per suite & 94.6 & 94.0 & 62.8 & 16.2 & 10.0 \\
\rowcolor{teal!7}  & ECT data & Four suites & 94.8 & 92.6 & 69.2 & 17.4 & 10.6 \\
\midrule
Long & Official & Per suite & 89.0 & 88.4 & 54.6 & 0.2 & 10.2 \\
\rowcolor{teal!7}  & ECT data & Per suite & 91.8 & 88.4 & 51.6 & 9.6 & 10.4 \\
\rowcolor{teal!7}  & ECT data & Four suites & 89.0 & 89.2 & 55.0 & 16.2 & 13.2 \\
\bottomrule
\end{tabular}
\end{table}

Over the official GR00T-N1.7, the per-suite ECT models raise Swap on Spatial, Goal, and Long, and the four-suite ECT data model raises it further on every suite while staying within 3 points on ID and lowering Task on no suite (\Cref{tab:app_groot_ect_full}). This model is compared with the official per-suite checkpoints, so part of its gain may come from training on four suites. Due to our computation budget, the per-suite ECT models are trained on roughly one-thirteenth of the samples of the official recipe, so the per-suite comparison favors the official checkpoints. Together, these results support applying the ECT data to a different action architecture.

The four-suite GR00T model remains below the Frozen-LM $\pi_{0.5}$ ECT model on Swap in all suites (\Cref{tab:app_pi05_unified_ect}). The recipes also differ in which modules are trained, with GR00T freezing its vision encoder and $\pi_{0.5}$ training it.

\subsection{Decomposition across seeds and suites}
\label{app:ect_decomposition}

The three $\pi_{0.5}$ models of \Cref{tab:ect_decomposition} (Standard, ECT data, and ECT data + ECT loss) share the Frozen-LM recipe, optimizer, 30k training steps, and 64 examples per update. The ECT loss fills each update with 32 counterpart pairs, and the other two models draw 64 independent examples.

\paragraph{Seed variation.}
All three models have three training seeds each (0, 1, 42). Across seeds, the largest Swap standard deviation is 4.9 points for ECT data + ECT loss (Long in \Cref{tab:pi05_expertonly_seed_values}), 6.9 for ECT data (Long), and 7.3 for Standard (Object).

\paragraph{Per-suite interpretation.}
The data component accounts for most of the Swap improvement on every suite, and pairing adds a further 2 to 5 points on Spatial, Goal, and Long. On Object, the ECT data alone already reach the Swap success of the full method (\Cref{tab:app_pi05_unified_ect}). On Task, the ECT loss has no consistent effect (\Cref{tab:app_pi05_unified_ect}). Its pairs share an instruction and differ only in the scene, and the Task gains over Standard come from the ECT data.

\subsection{Data-scale sweep}
\label{app:sweep}
\Cref{tab:sweep} tests whether the counterfactual gap is a matter of sample size (prediction (vi)). The Standard model is trained on 25\%, 50\%, and 100\% of the original four-suite demonstrations (subsampled per task) with the same base model, Frozen-LM recipe, and 30k steps, one training seed per point, and $N=500$ episodes per cell. Every Swap and Task cell stays at least 34.6 points below ID, and from 25\% to 100\% of the data Swap barely moves on three suites and falls on Object.

\begin{table}[htbp]
\centering
\scriptsize
\setlength{\tabcolsep}{3.5pt}
\caption{\textbf{More demonstrations of the same kind do not close the gap (prediction (vi)).} The Standard model (Frozen-LM recipe, 32 examples per update) trained on 25\%, 50\%, and 100\% of the original demonstrations, with one training seed and $N=500$ rollouts per cell.}
\label{tab:sweep}
\begin{tabular}{@{}l ccc ccc ccc@{}}
\toprule
 & \multicolumn{3}{c}{ID} & \multicolumn{3}{c}{Swap} & \multicolumn{3}{c}{Task} \\
\cmidrule(lr){2-4}\cmidrule(lr){5-7}\cmidrule(l){8-10}
Suite & 25\% & 50\% & 100\% & 25\% & 50\% & 100\% & 25\% & 50\% & 100\% \\
\midrule
Spatial & 95.2 & 95.8 & 90.0 & 53.6 & 57.4 & 55.4 & 22.6 & 26.8 & 21.2 \\
Object  & 98.2 & 98.6 & 97.8 & 39.6 & 35.8 & 23.8 & 20.2 & 36.4 & 16.6 \\
Goal    & 90.6 & 93.4 & 93.6 & 36.0 & 40.8 & 40.2 & 16.2 & 20.6 & 19.4 \\
Long      & 94.0 & 90.4 & 94.4 & 14.4 & 15.2 & 15.2 & 16.2 & 24.0 & 15.0 \\
\bottomrule
\end{tabular}
\end{table}

\subsection{ECT under RL post-training}
\label{app:pirl}
$\pi_{\mathrm{RL}}$ \citep{chen2025pi_} adds a standard RL stage after imitation, fine-tuning $\pi_{0.5}$ with PPO in LIBERO, rewarding success on the training instructions and scenes. It uses the original task instructions and scenes, but can collect new action trajectories. We test whether this standard post-training stage reduces the counterfactual gap and whether ECT's gains persist.

We run the RLinf implementation of $\pi_{\mathrm{RL}}$ with its stock PPO recipe for $\pi_{0.5}$ (64 environments, 8 rollout epochs of 240 steps per update, global batch 2048, 3 flow steps with stochastic sampling) on three four-suite Frozen-LM models: Standard and ECT data trained with 32 examples per update (pre-RL Standard is the 100\% model of \Cref{tab:sweep}), and the seed-42 ECT data + ECT loss checkpoint (\Cref{tab:pi05_expertonly_seed_values}). Each arm runs 40 PPO epochs with one seed. Training-time success rises from about 80\% to 88\% for all three. The RL checkpoints are converted back to JAX (a conversion that reproduces the original policy) and evaluated with the main LIBERO-PRO protocol.

After this RL stage, Standard remains within 4 points of its initialization on every Swap and Task cell (\Cref{tab:pirl}). Both ECT models still lead Standard on Swap in every suite and on Task in all but one cell, an exception inherited from the ECT data + ECT loss initialization seed (\Cref{tab:pi05_expertonly_seed_values}). The nuisance cells remain on par with Standard. RL lowers Swap on Long for both ECT models, narrowing their gains. These results describe 40 PPO epochs with one seed per arm.

\begin{table}[htbp]
\centering
\scriptsize
\setlength{\tabcolsep}{4pt}
\caption{\textbf{ECT gains survive RL post-training.} Frozen-LM Standard, ECT data, and ECT data + ECT loss models after 40 PPO epochs of $\pi_{\mathrm{RL}}$ (RLinf stock recipe, one seed). $N=500$ per cell. ECT rows are shaded, with the best result per column within each suite in bold.}
\label{tab:pirl}
\begin{tabular}{@{}llrrrrr@{}}
\toprule
Suite & Model & ID & Sem. & Obj. & Swap & Task \\
\midrule
Spatial & Standard, then $\pi_{\mathrm{RL}}$ & 96.0 & 92.0 & 93.0 & 55.4 & 20.2 \\
\rowcolor{teal!7} Spatial & ECT data, then $\pi_{\mathrm{RL}}$ & 95.0 & 91.4 & 94.4 & 69.4 & \textbf{28.2} \\
\rowcolor{teal!7} Spatial & ECT data + ECT loss, then $\pi_{\mathrm{RL}}$ & \textbf{96.4} & \textbf{94.0} & \textbf{95.0} & \textbf{75.0} & 15.2 \\
\midrule
Object & Standard, then $\pi_{\mathrm{RL}}$ & 98.0 & 98.0 & 83.0 & 22.4 & 16.8 \\
\rowcolor{teal!7} Object & ECT data, then $\pi_{\mathrm{RL}}$ & 99.2 & \textbf{99.4} & \textbf{89.0} & \textbf{68.6} & \textbf{39.0} \\
\rowcolor{teal!7} Object & ECT data + ECT loss, then $\pi_{\mathrm{RL}}$ & \textbf{99.4} & 99.0 & 81.0 & 68.4 & 30.6 \\
\midrule
Goal & Standard, then $\pi_{\mathrm{RL}}$ & 93.0 & 87.4 & 81.2 & 40.4 & 16.4 \\
\rowcolor{teal!7} Goal & ECT data, then $\pi_{\mathrm{RL}}$ & 91.8 & \textbf{91.2} & 84.6 & 45.8 & 31.4 \\
\rowcolor{teal!7} Goal & ECT data + ECT loss, then $\pi_{\mathrm{RL}}$ & \textbf{95.0} & 90.8 & \textbf{87.0} & \textbf{54.6} & \textbf{38.0} \\
\midrule
Long & Standard, then $\pi_{\mathrm{RL}}$ & 91.8 & \textbf{96.2} & 58.2 & 16.0 & 18.8 \\
\rowcolor{teal!7} Long & ECT data, then $\pi_{\mathrm{RL}}$ & \textbf{94.2} & 94.2 & 59.6 & \textbf{27.0} & 22.8 \\
\rowcolor{teal!7} Long & ECT data + ECT loss, then $\pi_{\mathrm{RL}}$ & 91.0 & 91.4 & \textbf{61.2} & 23.6 & \textbf{27.8} \\
\bottomrule
\end{tabular}
\end{table}

\section{ECT Beyond LIBERO}
\label{app:ect_beyond_libero}

This appendix details the UR5e study and CALVIN ABC$\to$D of \Cref{tab:beyond_libero}.

\subsection{Real-robot UR5e study}
\label{app:ur5}

\paragraph{Setup.} We use a UR5e arm with an OnRobot 2FG7 gripper, a fixed scene camera, and a wrist camera (both RGB). Demonstrations are collected by hand-guiding the arm. The training instruction is ``pick up the [object] and put it in the box'', and success means the object ends up in the box.
\paragraph{Scenes.} A cardboard box, which serves as the container, and four objects (a coke can, a coffee can, a milk candy, and a purple cube) stand on a table in front of the arm. The original layout places them at fixed positions. The three counterfactual layouts mirror this arrangement front to back (x-mirror), left to right (y-mirror), or both (xy-mirror), so the same instruction requires a different reach in each. All four layouts are set up physically and demonstrated by hand. 
\begin{figure}[H]
\centering
\includegraphics[width=\linewidth]{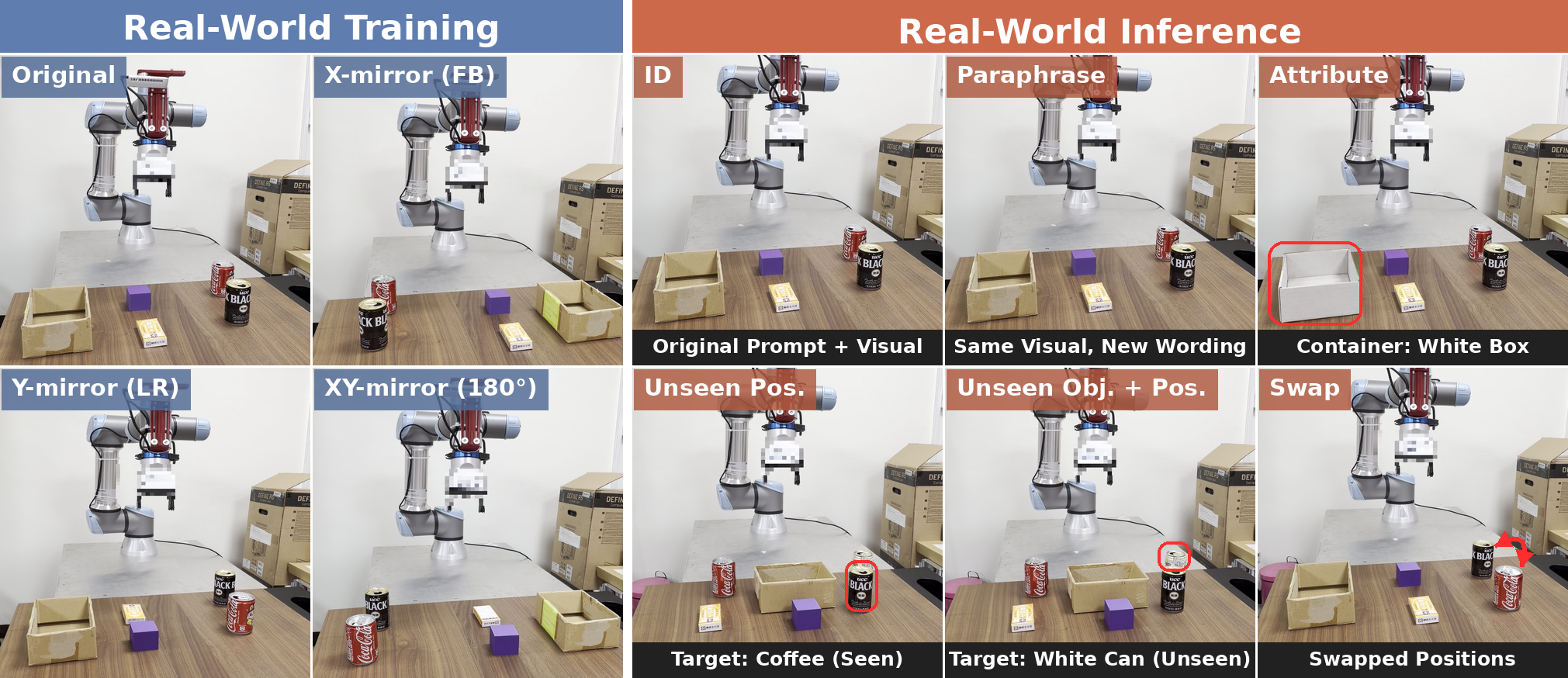}
\caption{\textbf{Real-world training layouts and evaluation settings.} The left block shows the original layout and three physically collected mirrored layouts. The right block illustrates the ID, paraphrase, attribute, unseen-position, unseen-object-and-position, and Swap conditions. The Swap comparison uses companion models trained without the layout that coincides with the swapped configuration.}
\label{fig:ur5_settings}
\end{figure}

\paragraph{Data budgets.} All three models are trained from the same $\pi_{0.5}$ base with the same recipe on 400 demonstrations. Standard uses 100 per object in a single layout, while each ECT model uses 25 per object in each of the four layouts. ECT data samples demonstrations independently. ECT data + ECT loss pairs each demonstration with one of the same object and grid position from a different layout, aligned by a linear time warp because human demonstrations are never frame-aligned.
\paragraph{Evaluation cells.} Each cell changes one factor relative to the original setting, except Unseen obj.\,+\,pos., which changes two. \emph{ID} uses the original layout, ten trials per object. \emph{Paraphrase} (nuisance) rewords the instruction as ``pick the [object] and place it into the box'', five trials per object. \emph{Attribute} (nuisance) changes an appearance at the trained positions under the unchanged instruction. A white coffee can replaces the coffee can, or a white box replaces the box, five trials each. \emph{Unseen position} (counterfactual) places the four objects and the box at positions that appear in no training layout, ten trials per object. \emph{Unseen object + position} (counterfactual) places the white coffee can, which never appears in training, at such positions, ten trials. \emph{Swap} (counterfactual) exchanges the coffee can and the coke can in the original layout under the unchanged instruction, with either can as the target. For this cell, each ECT model is replaced by a companion trained with the same method but without the layout that coincides with the swapped configuration, so that configuration is unseen by every evaluated model. All recorded trials are reported, ten per target can except for the ECT-data companion (five with the coffee can and nine with the coke can), and partial grasps count as failures. A Standard model trained on a quarter of the data (25 demonstrations per object, single layout) is as low on the unseen-position cells as the full one (1 against 3 of 50 trials), so four times more single-layout data do not move the counterfactual cells (prediction (vi)). The offline instruction-versus-vision cache decomposition is in \Cref{app:kv_patching}.

\subsection{CALVIN \texorpdfstring{ABC$\to$D}{ABC to D}}
\label{app:calvin}
CALVIN scenes A, B, and C place the same articulated fixtures (drawer, slider, switch, and button) at different positions, providing same-instruction demonstrations with different scene-dependent actions. All CALVIN models are $\pi_{0.5}$, trained with one seed per row, and evaluated on the official ABC$\to$D split with 1{,}000 sequences. \Cref{tab:beyond_libero} compares Standard training with pairing in the original data, and with constructed ECT data sampled independently or in pairs.

For original-data pairing, a demonstration is paired with a same-instruction demonstration from a different scene. For constructed data, a mirror reflects the whole table, including fixtures, and a shift moves selected fixtures or blocks. Each transformed demonstration is replayed, and only replays that pass CALVIN's task-success check are retained with their originals. The constructed-data rows are compute-matched, with about 0.27 epochs over the enlarged pool versus 1.5 epochs over the original data. The natural-pair result shows that additional constructed examples are not required for the reported gain. The comparison with construction also reflects these different coverage budgets.

\section{Statistical Uncertainty and Stability}
\label{app:uncertainty}

Single-checkpoint success rates ($N=500$ rollouts) and human-label proportions use Wilson confidence intervals, and differences use Newcombe-Wilson intervals. These capture episode-level uncertainty for a fixed checkpoint. Frozen-LM ECT data + ECT loss was trained with three independent seeds (0, 1, 42), so its primary stability summary is the seed mean and sample standard deviation. Hierarchical-bootstrap intervals over seeds and rollouts are descriptive given only three seeds.

\paragraph{Seed-level stability of ECT data + ECT loss.}
The Swap gains are stable across seeds on Spatial, Object, and Goal (\Cref{tab:pi05_expertonly_seed_values}). Long varies more, but every seed stays well above Standard (at least 31.6 against 12.7), so the gain over Standard appears in all three evaluated seeds.

\begin{table}[htbp]
\centering
\scriptsize
\setlength{\tabcolsep}{4pt}
\caption{\textbf{Seed-level success rates of $\pi_{0.5}$ ECT data + ECT loss (Frozen LM).}
Std.\ is the sample standard deviation across three independent training seeds.}
\label{tab:pi05_expertonly_seed_values}
\begin{tabular}{llccccc}
\toprule
Suite & Cell & Seed 0 & Seed 1 & Seed 42 & Mean & Std. \\
\midrule
Spatial & ID & 98.2 & 99.0 & 98.0 & 98.4 & 0.5 \\
Spatial & Sem. & 97.0 & 96.0 & 95.2 & 96.1 & 0.9 \\
Spatial & Obj. & 97.2 & 96.6 & 97.6 & 97.1 & 0.5 \\
Spatial & Swap & 75.4 & 75.0 & 72.8 & 74.4 & 1.4 \\
Spatial & Task & 21.8 & 28.8 & 15.6 & 22.1 & 6.6 \\
\midrule
Object & ID & 99.0 & 99.6 & 98.8 & 99.1 & 0.4 \\
Object & Sem. & 99.0 & 100.0 & 98.6 & 99.2 & 0.7 \\
Object & Obj. & 76.0 & 77.2 & 82.2 & 78.5 & 3.3 \\
Object & Swap & 74.8 & 70.2 & 67.4 & 70.8 & 3.7 \\
Object & Task & 38.4 & 26.0 & 29.6 & 31.3 & 6.4 \\
\midrule
Goal & ID & 94.4 & 96.6 & 95.8 & 95.6 & 1.1 \\
Goal & Sem. & 92.8 & 93.4 & 93.6 & 93.3 & 0.4 \\
Goal & Obj. & 85.8 & 88.0 & 86.2 & 86.7 & 1.2 \\
Goal & Swap & 55.8 & 55.8 & 55.0 & 55.5 & 0.5 \\
Goal & Task & 32.0 & 34.2 & 40.8 & 35.7 & 4.6 \\
\midrule
Long & ID & 92.8 & 93.4 & 92.2 & 92.8 & 0.6 \\
Long & Sem. & 94.0 & 92.4 & 92.2 & 92.9 & 1.0 \\
Long & Obj. & 57.6 & 58.8 & 60.8 & 59.1 & 1.6 \\
Long & Swap & 40.4 & 32.2 & 31.6 & 34.7 & 4.9 \\
Long & Task & 20.4 & 20.4 & 26.6 & 22.5 & 3.6 \\
\bottomrule
\end{tabular}
\end{table}

\paragraph{Success-rate deltas.}
Beyond the matched controls of \Cref{tab:counterfactual_main}, \Cref{tab:key_delta_uncertainty} compares Frozen-LM ECT data + ECT loss with the official $\pi_{0.5}$, a reference with more trainable parameters. For $\pi_{0.5}$, the Swap deltas are positive on every suite with intervals excluding zero, while ID stays within about one point. For GR00T-N1.7, the per-suite models leave Object Swap unchanged and lower ID on Spatial.

\begin{table}[htbp]
\centering
\scriptsize
\setlength{\tabcolsep}{3pt}
\caption{\textbf{Key ID and Swap deltas.}
$\pi_{0.5}$ intervals use a hierarchical bootstrap over seeds and rollouts and are descriptive because only three independent seeds are available. GR00T-N1.7 intervals are Newcombe hybrid-score intervals for a difference of two proportions ($N=500$ each).}
\label{tab:key_delta_uncertainty}
\begin{tabular}{lllcc}
\toprule
Model & Suite & Comparison & $\Delta$ Swap [95\% CI] & $\Delta$ ID [95\% CI] \\
\midrule
$\pi_{0.5}$ & Spatial & ECT data + ECT loss (Frozen LM) vs.\ official & $+27.8$ [+22.7, +32.7] & $0.0$ [-1.3, +1.4] \\
$\pi_{0.5}$ & Object & ECT data + ECT loss (Frozen LM) vs.\ official & $+52.6$ [+47.3, +57.9] & $+0.5$ [-0.6, +1.8] \\
$\pi_{0.5}$ & Goal & ECT data + ECT loss (Frozen LM) vs.\ official & $+21.3$ [+16.5, +26.2] & $-0.8$ [-2.9, +1.4] \\
$\pi_{0.5}$ & Long & ECT data + ECT loss (Frozen LM) vs.\ official & $+25.3$ [+20.0, +31.3] & $+1.0$ [-1.7, +3.9] \\
\midrule
GR00T-N1.7 & Spatial & ECT data, per suite vs.\ official & $+16.6$ [+13.2, +20.3] & $-6.6$ [-10.3, -2.9] \\
GR00T-N1.7 & Object & ECT data, per suite vs.\ official & $0.0$ [-0.8, +0.8] & $+2.0$ [-0.3, +4.4] \\
GR00T-N1.7 & Goal & ECT data, per suite vs.\ official & $+14.0$ [+10.6, +17.6] & $+0.4$ [-2.5, +3.3] \\
GR00T-N1.7 & Long & ECT data, per suite vs.\ official & $+9.4$ [+6.9, +12.3] & $+2.8$ [-0.9, +6.5] \\
\bottomrule
\end{tabular}
\end{table}

\paragraph{Human-label proportion uncertainty.}
Wilson intervals for the human-label proportions of \Cref{sec:human_rollout} (\Cref{tab:human_label_uncertainty}) capture finite-sample uncertainty only. \Cref{tab:annotation_reliability} reports inter-annotator agreement.

\begin{table}[htbp]
\centering
\scriptsize
\setlength{\tabcolsep}{3pt}
\caption{\textbf{Wilson confidence intervals for human rollout-label proportions.}
Retrieval-like includes source retrieval, other-task retrieval, and hybrid retrieval. Counts follow \Cref{tab:app_human_rule_counts}.}
\label{tab:human_label_uncertainty}
\begin{tabular}{lccc}
\toprule
Scope & Retrieval-like & Correct/near-correct & Collapse \\
\midrule
  All probes & 264/381 ($69.3$ [64.5, 73.7]) & 101/381 ($26.5$ [22.3, 31.2]) & 16/381 ($4.2$ [2.6, 6.7]) \\
  Goal & 99/120 ($82.5$ [74.7, 88.3]) & 15/120 ($12.5$ [7.7, 19.6]) & 6/120 ($5.0$ [2.3, 10.5]) \\
  Long & 63/72 ($87.5$ [77.9, 93.3]) & 7/72 ($9.7$ [4.8, 18.7]) & 2/72 ($2.8$ [0.8, 9.6]) \\
  Object & 40/69 ($58.0$ [46.2, 68.9]) & 28/69 ($40.6$ [29.8, 52.4]) & 1/69 ($1.4$ [0.3, 7.8]) \\
  Spatial & 62/120 ($51.7$ [42.8, 60.4]) & 51/120 ($42.5$ [34.0, 51.4]) & 7/120 ($5.8$ [2.9, 11.6]) \\
\bottomrule
\end{tabular}
\end{table}

\paragraph{Prefix-KV uncertainty.}
Prefix-KV uncertainty appears with the diagnostics in \Cref{app:kv_patching}. Retrieval rates are exact counts over the 45 \texttt{same\_to\_other} rollouts, and patching estimates use bootstrap intervals over rollouts. For target specificity, candidate-level metrics are aggregated per rollout before the bootstrap, so the 450 row-candidate pairs are not treated as independent.

\section{Relation to Other Accounts of VLA Failure}
\label{app:other_accounts}
Binding characterizes the structured failures in \Cref{sec:human_rollout}. Work on vision shortcuts and modality imbalance \citep{lian2026bayesianvla,fei2025libero,fang2026vision,xu2025seeing,darabi2026progal} motivates testing how language influences action selection. On Goal, altered instructions select other demonstrated behaviors, and the scene-matched prefix readout recovers the executed task in $34/45$ cases, versus $22/45$ for surface wording. Language therefore affects which familiar behavior is selected in these probes, even when that behavior does not satisfy the request.

The Object stress test isolates a different aspect of the failure. The policy tracks a displaced object after selecting the wrong target, demonstrating retained visual feedback during execution. This observation concerns the use of visual information and does not exclude changes to the visual representation during fine-tuning \citep{kachaev2025don}. All $\pi_{0.5}$ recipes train the vision encoder (\Cref{app:recipes}). Together, the source-task, other-task, and hybrid outcomes distinguish familiar-behavior selection from unstructured collapse \citep{metz2016unrolled,chi2025diffusion}.

\section{Limitations}
\label{app:limitations}
Swap and Task remain below ID after ECT in every suite (\Cref{tab:app_pi05_unified_ect}). Binding characterizes task selection rather than a unique retrieval algorithm. Pairing is not uniformly beneficial. In $\pi_{0.5}$ the paired sampler changes counterpart co-presentation and the balance of original and counterpart examples together, and the two halves share flow-matching noise and time, so $\Delta_{\mathrm{loss}}$ measures these ingredients jointly. Its optimization mechanism remains open.

The full $\pi_{0.5}$ method and its Frozen-LM comparisons use three training seeds, whereas CALVIN and several other comparisons use one. Rollout intervals exclude training variability. Official checkpoints are not matched retraining controls, and four-suite GR00T changes training scope. The data sweep covers a fourfold range at fixed steps.

ECT requires action-valid, scene-dependent alternatives. We test constructed, natural, and physically collected pairs after task-specific fine-tuning. Generalist use without such fine-tuning, contact-rich manipulation, mobile manipulation, and navigation remain untested.

\end{document}